\documentclass[conference]{IEEEtran}
\usepackage{times}

\usepackage[numbers]{natbib}
\usepackage{multicol}
\usepackage[bookmarks=true]{hyperref}

\usepackage{graphicx}
\usepackage{subcaption}
\usepackage{wrapfig}

\usepackage{array}
\usepackage{booktabs}
\usepackage{multirow}

\usepackage{amsmath,amsfonts,amssymb,amsthm}
\usepackage{mathtools}
\usepackage{commath}

\usepackage{xcolor}
\definecolor{color_train}{rgb}{0.415686275, 0.658823529, 0.309803922}
\definecolor{color_test}{rgb}{0.403921569, 0.305882353, 0.654901961}
\definecolor{lightgrey}{rgb}{0.43,0.43,0.43}
\definecolor{deepmagenta}{rgb}{0.8, 0.0, 0.8}
\definecolor{yes_color}{rgb}{0.0, 0.6, 0.0}

\usepackage{algorithm}
\usepackage{algpseudocode}

\begin{document}

% paper title
\title{Factored World Models for Zero-Shot \\Generalization in Robotic Manipulation}

%\author{Author Names Omitted for Anonymous Review. Paper-ID 99}

\author{\authorblockN{Ondrej Biza$^1$, Thomas Kipf$^3$, David Klee$^1$, Robert Platt$^1$, \\Jan-Willem van de Meent$^{1,2}$ and Lawson L. S. Wong$^1$\vspace{1em}}
\authorblockA{${^1}$Khoury College of Computer Sciences, Northeastern University \\
${^2}$Informatics Institute, University of Amsterdam \\ 
${^3}$Google Research, Brain Team \\ 
\texttt{Email: biza.o@northeastern.edu}}}

% \author{\authorblockN{Ondrej Biza}
% \authorblockA{Northeastern University\\
% biza.o@northeastern.edu}
% \and
% \authorblockN{Thomas Kipf}
% \authorblockA{Google Research\\
% tkipf@google.com}
% \and
% \authorblockN{David Klee}
% \authorblockA{Northeastern University\\
% klee.d@northeastern.edu}
% \and
% \authorblockN{Robert Platt}
% \authorblockA{Northeastern University\\
% rplatt@ccs.neu.edu}
% \and
% \authorblockN{Jan-Willem van de Meent}
% \authorblockA{University of Amsterdam\\
% Northeastern University\\
% j.w.vandemeent@uva.nl}
% \and
% \authorblockN{Lawson L. S. Wong}
% \authorblockA{Northeastern University\\
% lsw@ccs.neu.edu}}

\maketitle

\begin{abstract}
World models for environments with many objects face a combinatorial explosion of states: as the number of objects increases, the number of possible arrangements grows exponentially. In this paper, we learn to generalize over robotic pick-and-place tasks using object-factored world models, which combat the combinatorial explosion by ensuring that predictions are equivariant to permutations of objects. Previous object-factored models were limited either by their inability to model actions, or by their inability to plan for complex manipulation tasks. We build on recent contrastive methods for training object-factored world models, which we extend to model continuous robot actions and to accurately predict the physics of robotic pick-and-place. To do so, we use a residual stack of graph neural networks that receive action information at multiple levels in both their node and edge neural networks. Crucially, our learned model can make predictions about tasks not represented in the training data. That is, we demonstrate successful zero-shot generalization to novel tasks, with only a minor decrease in model performance. Moreover, we show that an ensemble of our models can be used to plan for tasks involving up to 12 pick and place actions using heuristic search.  We also demonstrate transfer to a physical robot.
\end{abstract}

\IEEEpeerreviewmaketitle

\section{Introduction}
\label{sec:introduction}

From assembly to household robots, current state-of-the-art robot learning agents cannot generalize beyond a specific training task. One important aspect of generalization is the ability to understand any novel combination of known factors, a so-called combinatorial or compositional generalization. Applied to objects, combinatorial generalization ideally allows an agent to understand any arrangement of objects from only a limited number of interactions with its environment. The two key steps such agents need to perform are (a) decomposing a scene into individual objects and (b) modeling relative interactions between objects. The first step can be implemented using any object recognition model from computer vision \cite{zhao19object}. The modeling of object interactions is particularly important when learning to predict the dynamics of the environment (i.e. learning a world model).

The purpose of our paper is to learn a world model that accurately predicts the effects of actions in the context of robotic manipulation, and generalizes to novel tasks. To this end, we propose factored world models (FWMs), which build on recently-proposed contrastive methods \cite{kipf20contrastive} to represent continuous robot actions and predict outcomes of pick-and-place tasks in domains with realistic physical interactions. In FWMs, we decouple the object recognition from world modeling by assuming an input state that is already factored into individual images for each object, and focus on learning a latent code for objects and predicting the effect of pick-and-place actions. FWMs are trained using a self-supervised contrastive loss. They can predict accurate manipulation physics by using a stack of multiple graph neural network layers (GNNs) with residual connections. GNNs consider pairwise interactions, usually between every possible pair of objects, in order to predict the state of each object one step into the future. GNNs achieve combinatorial generalization (to the extent that pairwise interactions are expressive enough) by equivariance to the order in which objects are presented--permutation equivariance. That is, they cannot overfit to a particular ordering of objects. It is a common choice to use a single GNN layer (in some cases called an Interaction Network \citep{battaglia16interaction}). But, as shown in previous work \citep{kipf2018neural,sanchez2018graph} and confirmed by our experiments, stacking multiple GNN layers leads to much more accurate physics predictions.

\begin{figure}[t]
  \centering
  \includegraphics[width=0.8\linewidth]{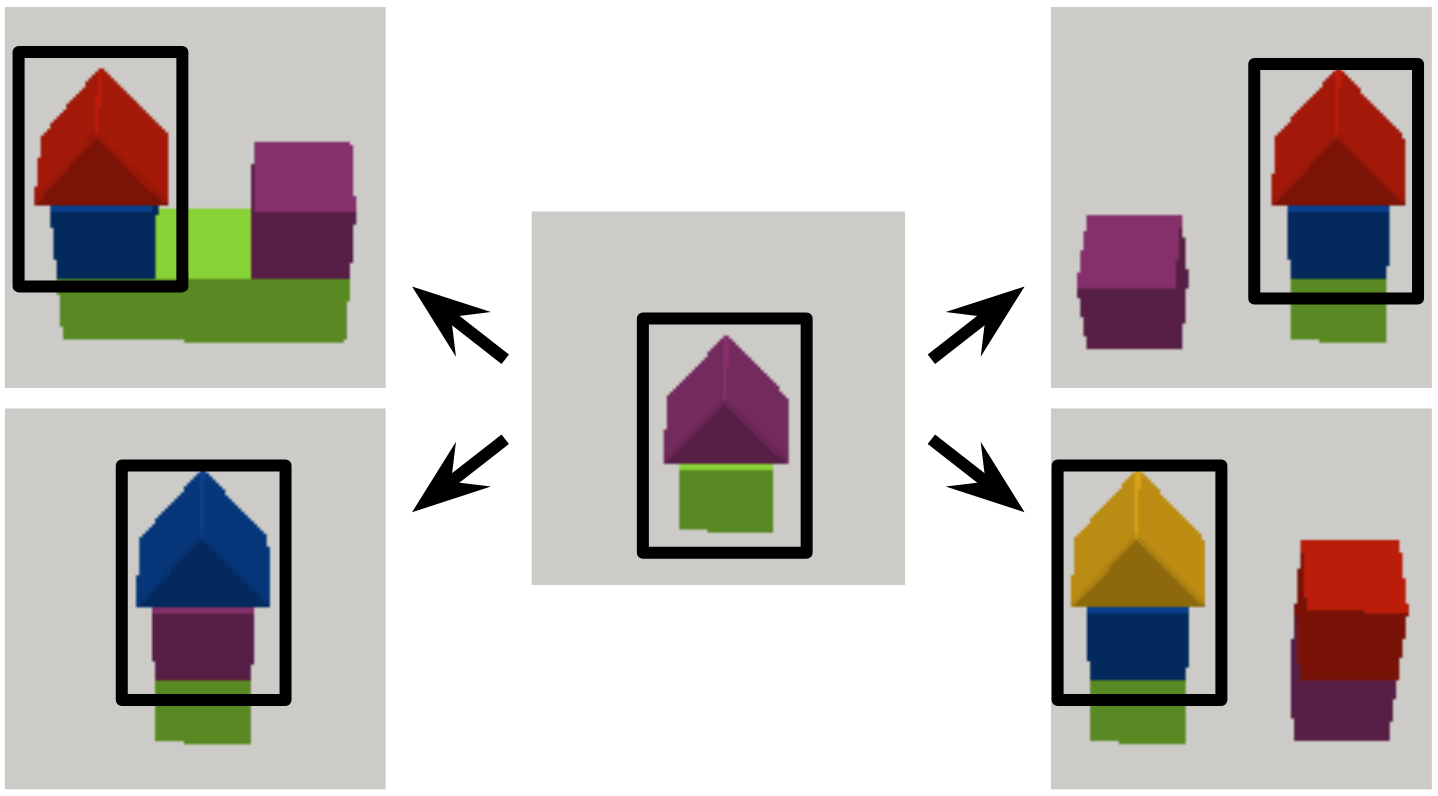}
  \caption{By considering pairwise interactions of objects, a graph neural network can generalize a substructure ("a triangle on top of a block" in this case) to unseen tasks.\vspace{-1em}}
  \label{fig:gen_example}
\end{figure}

The ability to integrate information about actions with latent states of individual objects has been underexplored in prior work. Previous object-factored world models either assume object-action association (a factored action space) \citep{kipf20contrastive,huang20better} or only model sequences of states without actions \citep{janner19reasoning,bakhtin19phyre,qi21learning}. We represent actions as \texttt{pick(x,y)} and \texttt{place(x,y)} with continuous \texttt{(x,y)} coordinates; this representation is integrated into our latent transition model through an iterative refinement by a residual stack of graph neural network layers. At each level of the stack, the action coordinates are injected both into the node and edge network of the GNN layer. For example, the GNN is able to model the effect of an incorrect placement that knocks over a tower by considering how an action affect all blocks in the structure. \citep{veerapaneni19entity} proposed a model that integrates the \texttt{(x,y)} action coordinates, but their pick-and-place tasks are limited to between two and four time steps. FWMs learn tasks involving up to eight objects and twelve pick-and-place actions. 

\begin{figure}[t]
  \centering
  \includegraphics[width=0.5\linewidth]{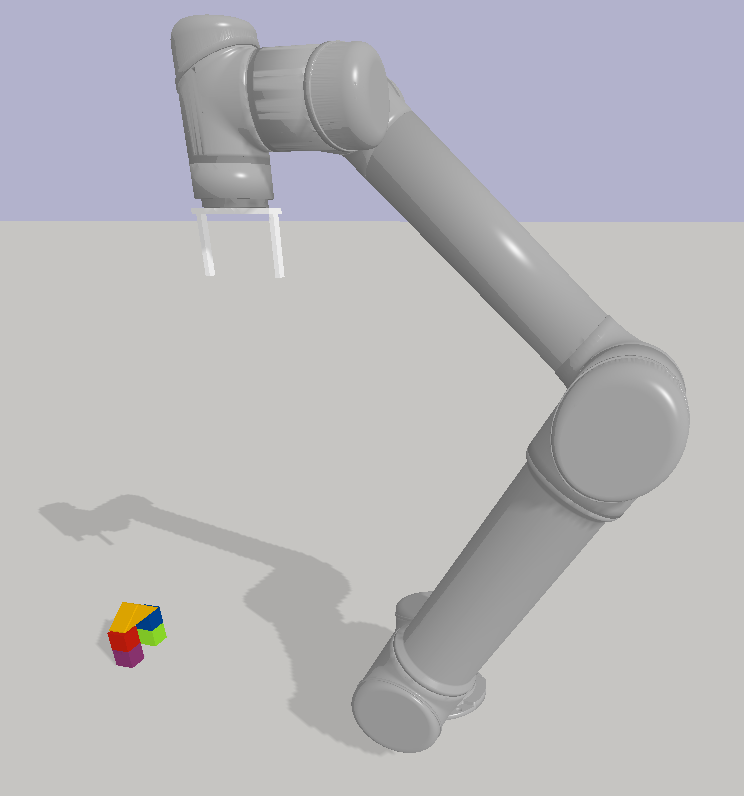}
  \caption{A simulated block pick-and-place tasks with a UR5 robotic arm and a antipodal gripper.}
  \label{fig:envs}
\end{figure}

% environments
We train and evaluate FWMs in a simulated environment involving a UR5 robotic arm manipulating blocks of various shapes (Figure \ref{fig:envs}) as well as on a physical robot (Figure \ref{fig:real_env}). We instantiate two environments: \textsc{Cubes} includes six cubes arranged into five different structures, which take up to ten actions to build, and \textsc{Shapes} includes eight blocks of four types (cube, brick, triangle, roof), which are arranged into sixteen different structures (Table \ref{tab:tasks}).

To demonstrate zero-shot generalization --- the main result of our paper --- we partition tasks into training and zero-shot transfer sets, which ensures that FWMs never sees the goal state of any transfer task during training. We show that FWMs transfer to the zero-shot tasks in \textsc{Cubes} and \textsc{Shapes} with only a very minor drop in accuracy (Table \ref{tab:main_comp}). The mechanism behind successful zero-shot transfer is the ability of FWMs to understand the state of each object independently of others, and the ability to model sparse interactions. For example, if our model understands the interaction "stack a triangle on top of a cube" based on the first training task in \textsc{Shapes} (Table \ref{tab:tasks}, bottom row), it can use it when generalizing to four different held out tasks (Figure \ref{fig:gen_example}).

We evaluate FWMs both offline and online. For offline evaluation, we design two different metrics based on block position prediction and action sequence ranking. The purpose of the metrics is to quickly predict the performance of the model as a proxy for downstream tasks. In online evaluation, we use one-step heuristic search with an ensemble of models. Our system can plan for tasks involving up to twelve pick and place actions with around 70\% success rate. These tasks have not been seen during training and we make plans for them based on only one example of the goal state. Finally, we demonstrate transfer of models trained in simulation to a physical robotic setup -- without fine-tuning -- with around 40\% success rate.\\

% contributions
\noindent In summary, we contribute the following:
\begin{enumerate}
    \item We develop a factored wold model for robotic pick-and-place tasks. This model can accurately model the effect of continuous \texttt{(x,y)} robot actions, owing to a residual stack of graph neural networks.
    \item We demonstrate that the learned world model accurately predicts the outcomes of sequences of actions. The model achieves zero-shot generalization to tasks and object configurations that were not seen during training, including sequences that are longer (up to 12 actions) than those in the training data (up to 8 actions).
    \item Using one-step heuristic search with an ensemble of factored world models, we demonstrate successful robot control both in simulation and on a physical robot.
\end{enumerate}

\begin{flushleft}
\begin{table*}[t!]
    \centering
    \begin{tabular}{lc@{\hskip3pt}c@{\hskip3pt}c@{\hskip3pt}c@{\hskip3pt}c@{\hskip3pt}c@{\hskip3pt}c@{\hskip3pt}cc@{\hskip3pt}c@{\hskip3pt}c@{\hskip3pt}c@{\hskip3pt}c@{\hskip3pt}c@{\hskip3pt}c@{\hskip3pt}c@{\hskip3pt}}
        \toprule
         & \multicolumn{8}{c}{\textbf{Training Tasks, Success Rate}} & \multicolumn{8}{c}{\textbf{Zero-Shot Transfer Tasks, Success Rate}} \\
        \midrule
        \rotatebox[origin=c]{90}{\ \ \textsc{Cubes}} & \includegraphics[width=0.05\textwidth]{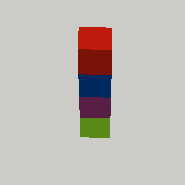} & & & & & & & & \includegraphics[width=0.05\textwidth]{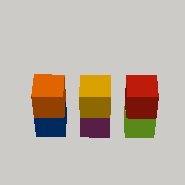} & \includegraphics[width=0.05\textwidth]{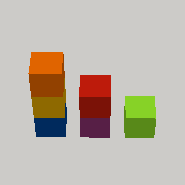} & \includegraphics[width=0.05\textwidth]{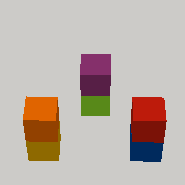} & \includegraphics[width=0.05\textwidth]{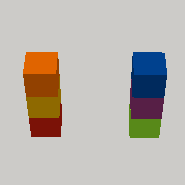}\vspace{-1.2em} & & & & \\
         & 95\% & & & & & & & & 70\% & 70\% & 70\% & 70\% & & & & \\
        \midrule
        \rotatebox[origin=c]{90}{\ \ \textsc{Shapes}} & \includegraphics[width=0.05\textwidth]{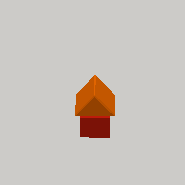} & \includegraphics[width=0.05\textwidth]{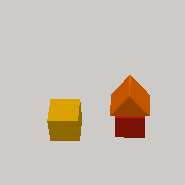} & \includegraphics[width=0.05\textwidth]{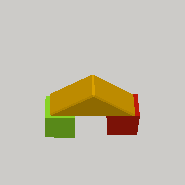} & \includegraphics[width=0.05\textwidth]{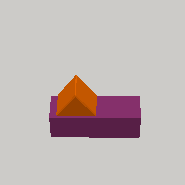} & \includegraphics[width=0.05\textwidth]{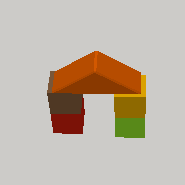} & \includegraphics[width=0.05\textwidth]{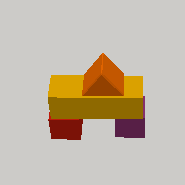} & \includegraphics[width=0.05\textwidth]{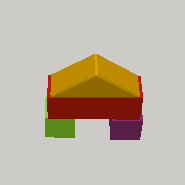} & \includegraphics[width=0.05\textwidth]{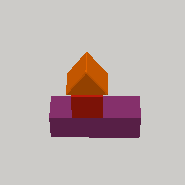} & \includegraphics[width=0.05\textwidth]{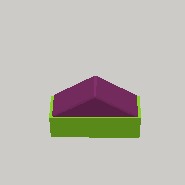} & \includegraphics[width=0.05\textwidth]{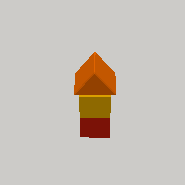} & \includegraphics[width=0.05\textwidth]{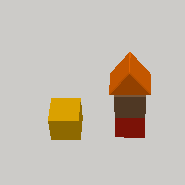} & \includegraphics[width=0.05\textwidth]{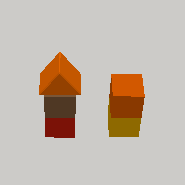} & \includegraphics[width=0.05\textwidth]{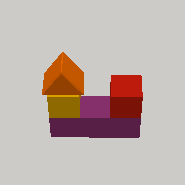} & \includegraphics[width=0.05\textwidth]{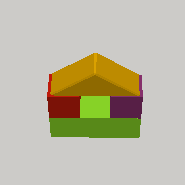} & \includegraphics[width=0.05\textwidth]{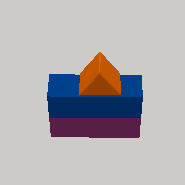} & \includegraphics[width=0.05\textwidth]{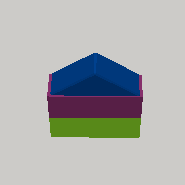}\vspace{-1.2em} \\
         & 85\% & 70\% & 70\% & 80\% & 55\% & 60\% & 75\% & 60\% & 100\% & 60\% & 90\% & 60\% & 30\% & 50\% & 65\% & 85\% \\ 
        \bottomrule
    \end{tabular}
    \caption{Examples of goals states of training and zero-shot transfer tasks for \textsc{Cubes} and \textsc{Shapes}. We collect expert demonstrations for the training tasks and test zero-shot generalization on the transfer tasks. The figure also includes a breakdown of success rates for heuristic search with an ensemble of 20 factored world models (Section \ref{sec:experiments:planning}, Table \ref{tab:planning}).}
    \label{tab:tasks}
\end{table*}
\end{flushleft}

\section{Related Work}

\textbf{World Models with Object Factorization.} Action-conditioned object-factorized world models have been explored in toy object environments and Atari games \citep{kipf20contrastive,huang20better} as well as in robotic manipulation \citep{veerapaneni19entity}, controlling billiard balls \citep{kossen20structured} and in navigation \citep{corneil18efficient}. We build upon C-SWM \citep{kipf20contrastive}, which has been further extended by \citep{huang20better} to handle identically looking objects (one of the assumptions of C-SWM is that each object is distinct). \citet{veerapaneni19entity} used a factored transition model to learn and plan for stacking up to three cubes. Different from these works we consider a simplified setup wherein our model does not perform symmetry breaking or object discovery from raw visual features, since we assume our environment to be already factored.

Object-factored world models are also commonly used in the context of physics prediction in videos. \citet{janner19reasoning,veerapaneni19entity} demonstrated generalization in a Tetris-like environment, where a world model predicts the outcome of dropping blocks from a height. \citet{ye19object} studied learning dynamics of pushing objects with a robotic arm in the real world. RPIN \citep{qi21learning} demonstrated state-of-the-art results in the tasks of 2D physics modeling (PHYRE benchmark, \cite{bakhtin19phyre}), modeling billiard balls and modeling videos of falling stacks of blocks. As we showed in our comparison, models like RPIN cannot be naively applied to pick-and-place tasks, both due to the problem of including action information at the right point in a transition model and due to the large changes between subsequent states in our environments (object disappearing from the environment vs.~a video of an object being slowly picked up by a robotic hand).

Object factorization is an instance of a general factored Markov Decision Process, which has been studied in the context of policy search \citep{guestrin03efficient}, factor discovery \citep{jonsson05casual} and MDP abstraction \citep{ravindran2004,wolfe06decision}. In fact, factored MDP works often include examples where individual factors are objects (e.g.~Tower of Hanoi in \citet{ravindran2004}). Early works specifically on object factorization include \citet{wolfe06defining,diuk08object}.

\textbf{Model-based Learning for Robotics.} One prominent line of work focuses on optimal control or policy search in low-dimensional state space. If such space is directly available, PILCO fits a Gaussian Process to real-world robot dynamics and performs policy search through gradient descent in the model \citep{deisenroth11pilco}. Successful cart-pole swing and robotic unicycle balance is achieved within a few seconds of online learning. In the case of image states, a low-dimensional latent space can be learned by variational autoencoders (see \cite{lesort18state} for a survey). \citet{watter15embed} learn a latent state space together with a locally linear transition model, enabling efficient inference in optimal control methods. \citet{watter15embed} demonstrate successful control of pendulum, cart-pole and a simplified robotic arm operating in 2D. Follow-up works add additional constraints to the framework in order to improve dynamics model accuracy and robustness \citep{karl17deep,banijamali18robust,levine20prediction}.

Other works have used models of the world to accelerate learning for robotic manipulation tasks. In \cite{nair2018visual}, a VAE, trained on image observations, is used for sampling goals during training and providing a reward signal via distance in the latent space. This method achieved similar sample complexity to a state-based method on a robotic pick-and-place task.  In \cite{nasiriany2019planning}, a challenging robotic pushing task was performed by searching for sequences of subgoals in the latent space of a VAE that could be followed by a learned policy. In contrast, we model environments with six to eight objects and demonstrate generalization to novel tasks.

\section{Factored World Models (FWMs)}
\label{sec:methods}

World models predict the effect of a sequence of actions $a^1, a^2, ..., a^T$ in an environment that is initialized to a state $s^0$. We will assume a setting in which the state of the world is represented as an image, which has been pre-processed into a factorized state $s=\langle s_1, s_2, ..., s_K \rangle$ in which each $s_i$ is an image centered on the $i$th object. This postulated factorization can be implemented using an object detection module that predicts a bounding box for each object and a tracking module that corresponds bounding boxes across time. Both problems have been studied extensively in computer vision \citep{zhao19object,greff20binding}. 

FWMs learn a factorized representation $z = \langle z_1, z_2, ..., z_K \rangle$, along with a transition model that predicts the state $z'$ that results from taking an action $a$ at state $z$. We will show that the learned latent code is low-dimensional; it represents object identity and position while abstracting over irrelevant features such as object color. FWMs are implemented using a graph neural network \citep{gori05new} that represents action-conditioned pairwise interactions of latent factors. It is able to perform several rounds of message passing, each successively refining the prediction of the physics of picking and placing objects. We do not learn a mapping from $z$ to $s$ (i.e.~the world model only predicts the future within its latent code); nevertheless, we show the latent transition model successfully solves prediction and action ranking tasks. An ensemble of FWMs can be used to plan for novel tasks given a single example of a goal state.

Both the state encoder and the latent transition model are equivariant to the permutation of factors. That is, we can provide objects in any order as long as it is consistent within an episode. Moreover, the model is agnostic to the number of objects present in a scene -- no parameters in the model are dependent on the number of objects -- although the weights of the latent transition model might overfit if the number does not vary during training. We describe FWMs in detail in Section \ref{sec:methods:model} and the training procedure in Section \ref{sec:methods:learning}. Section \ref{sec:methods:search} covers planning with an ensemble of FWMs.

\subsection{Model Architecture}
\label{sec:methods:model}

\textbf{Encoder.} Given a factored state $s = \langle s_1, s_2, ..., s_k \rangle$, the encoder $f_\phi$ processes each factor (image of an object) separately. That is, $z = \langle f_\phi(s_1), f_\phi(s_2), ..., f_\phi(s_K) \rangle$. The encoder uses three convolutional layers followed by average pooling and three fully-connected layers. In principle, the use of average pooling enables the encoder to process images of any size, although we use a fixed image size. As described in Section \ref{sec:experiments}, the per-object image $s_i$ includes both an RGB component and a coordinate grid component, which marks the location where the crop was taken (Figure \ref{fig:observation}).

\textbf{Transition Model.} The transition model $g_\theta$ accepts a factored latent state $z = \langle z_1, z_2, ..., z_K \rangle$, $z_i \in \mathbb{R}^{D_z}$ and an action $a \in \mathbb{R}^{D_a}$. It predicts the next latent state by outputting a residual:
\begin{align}
    \hat{z}^{t + 1} = z^t + g_\theta(z^t, a^t).
\end{align}
We implement $g_\theta(z^t, a^t)$ as a stack of graph neural network layers with skip connections. Denoting the $i$th graph neural network layer as $\text{GNN}_i$ and the $i$th intermediate representation as $y_i$, the computation with $L$ layers is as follows:
\begin{align}
    y_1 &= z + \text{GNN}_1(z, a) \nonumber \\
    y_2 &= y_1 + \text{GNN}_2(y_1, a) \nonumber \\
    ... \nonumber \\
    g_\theta(z, a) &= y_{L - 1} + \text{GNN}_L(y_{L - 1}, a)
\end{align}
Our implementation of individual graph neural network layers follows \citet{kipf20contrastive}: the network models pairwise interactions between latent state factors (corresponding to objects in the environment) using a fully-connected node network $h_n$ and an edge network $h_e$. Both are implemented as MLPs with one hidden layer. The edge network outputs an embedding for each directed edge $e_{i,j} = h_e(z_i, z_j, a)$. The edge embeddings are then aggregated using the node network in order to update the state of each node. Finally, the graph neural network outputs a vector of updated factors $z' = \langle z'_1, z'_2, ..., z'_K \rangle$ with
\begin{align}
    z'_i = h_n \bigg(z_i, a, \sum_{j \neq i} e_{j, i} \bigg).
\end{align}
The summation over edges in the input of the node network assures permutation equivariance, an important property that aids generalization to novel combinations of factors. Given a permutation of factors $\pi$, we have
\begin{align}
    \text{GNN}(\pi(z), a) = \pi(\text{GNN}(z, a)).
\end{align}
The permutation equivariance property also holds for a stack of GNN layers as well as the state encoder.

\subsection{Learning by Contrastive Loss}
\label{sec:methods:learning}

We train the encoder and latent transition model using a single-step contrastive loss. We use a contrastive loss with a single positive and negative example \citep{kipf20contrastive}. Given a transition $\langle s^t, a^t, s^{t+1} \rangle$, we encode the current state $z^t$, the next state $z^{t+1}$ and predict the next state $\hat{z}^{t+1}$ from the current state,
\begin{align*}
    z^t &= f_\phi(s^t), &
    z^{t+1} &= f_\phi(s^{t+1}), &
    \hat{z}^{t+1} = z^t + g_\theta(z^t, a^t).
\end{align*}    
The contrastive loss minimizes the distance between the real and predicted next latent state, while maximizing the encoding distance between the current state and a negative state ($\bar{z} = f_\phi(\bar{s}))$ up to a margin $\gamma$:
\begin{align}
    &L(z^t, z^{t+1}, \hat{z}^{t+1}, \bar{z}) =  \frac{1}{2K\sigma^2} \sum_{i=1}^K \norm{z_i^{t+1} - \hat{z_i}^{t+1}}_2^2 \nonumber \\
    &+ \max \left\{ 0, \gamma - \frac{1}{2K\sigma^2} \sum_{i=1}^K \norm{z_i^t - \bar{z_i}}_2^2 \right\}.
\end{align}
The negative state $\bar{s}$ is sampled by randomly permuting a training batch of current states. It is a proxy for sampling a random state from the entire dataset (without needing to increase the batch size). Intuitively, we want the encoder to capture the minimum information required to distinguish a randomly sampled pair of states while enabling the latent transition model to be accurate.

\subsection{Planning for Novel Tasks}
\label{sec:methods:search}

\begin{algorithm}

\caption{Pick-and-Place Heuristic}\label{alg:heuristic} 

\begin{flushleft}
    \hspace*{\algorithmicindent} \textbf{Input:} State embedding $z$, goal embedding $z^g$, action $a$, transition model $g_{\theta}$, in-hand classifier $j_{\psi}$, goal threshold $\delta$. \\
    \hspace*{\algorithmicindent} \textbf{Output:} Heuristic distance. \\
\end{flushleft}

\begin{algorithmic}[1]

    \State Predict next state $\hat{z} = z + g_{\theta}(z, a)$.
    \State Heuristic $h \leftarrow 0$.
    \State Match objects between $\hat{z}$ and $z^g$ (Section \ref{sec:methods:search}).
    \For {$k$th object}
        \State $d \leftarrow \norm{\hat{z}_k - z^g_k}^2_2$
        \If {$d < \delta$}
            \State $h \leftarrow h + d$. \Comment{Close enough to goal.}
        \ElsIf{$j_{\psi}(\hat{z}_k) = 1$}
            \State $h \leftarrow h + 1$. \Comment{At least one step away from goal.}
        \Else
            \State $h \leftarrow h + 2$. \Comment{At least two steps away from goal.}
        \EndIf
    \EndFor
    \State \Return $h$.

\end{algorithmic}

\end{algorithm}

% FWM can be used to plan for novel tasks given a single example
% greedy search w/ heuristic, spatial action space, don't need CEM
Given a trained FWM, we evaluate our ability to plan for a new task based on a single example of its goal state. We perform planning in an online setting, where the agent selects a robot action at each time step. To successfully build a structure, the agent has to make correct decisions about picking and placing objects instead of pushing them around, and about building layers of a structure from the bottom up. This kind of reasoning could be performed by standard search algorithms such as Monte Carlo Tree Search \citep{sutton18reinforcement}, multi-step cross-entropy method \citep{mannor03cross,veerapaneni19entity} or gradient-based planning \citep{lavalle06planning}, but at a prohibitive cost. Since we use a spatial action space (\texttt{pick(x, y)} and \texttt{place(x, y)}), it is highly unlikely that a randomly sampled trajectory will execute a sequence of successful picks and places. Moreover, the space of trajectories is non-smooth, making it difficult to iteratively improve randomly sampled trajectories. A small perturbation of a pick or place coordinate might cause an action to fail, drastically changing the value of an entire trajectory.

% heuristic for interchangeable objects
% hungarian matching algorithm
% in-hand classifier
Our method of choice is a one-step heuristic search. While we are limited by the quality of the heuristic, one-step search allows us to explore the outcomes of ten thousand actions, sampled in a $100{\times}100$ grid spanning the workspace, in a hundred milliseconds. The planning time increases to about two seconds for an ensemble of 20 models. In contrast, any multi-step search would take at least several minutes per timestep in our setting. The heuristic encodes elementary information about pick-and-place tasks, such as the fact that picking and placing blocks is often more advantageous than pushing them to their goal positions. Note that the heuristic operates solely over the latent space of FWMs; it does not require the ground-truth state of the environment. We describe two versions of our heuristic below, which are used in \textsc{Cubes} and \textsc{Shapes} respectively.

\begin{table}[t!]
    \centering
    \begin{tabular}{lrrrr}
        \toprule
        \textbf{Method} & \multicolumn{2}{c}{\textbf{\textsc{Cubes}--Training}} & \multicolumn{2}{c}{\textbf{\textsc{Cubes}--Zero-Shot}} \\
         & \begin{tabular}{@{}c@{}} RMSE \\ (cm)\end{tabular} & \begin{tabular}{@{}c@{}} Hits@1 \\ (\%)\end{tabular} & \begin{tabular}{@{}c@{}} RMSE \\ (cm)\end{tabular} & \begin{tabular}{@{}c@{}} Hits@1 \\ (\%)\end{tabular} \\
        \midrule
        RPIN (\cite{qi21learning}) & 20.0 & 0 & 21.7 & 0 \\
        C-SWM (\cite{kipf20contrastive}) & 11.39{\color{lightgrey}\tiny$\pm$0.02} & {0}{\color{lightgrey}\tiny$\pm$0.0} & {9.15}{\color{lightgrey}\tiny$\pm$0.04} & {0}{\color{lightgrey}\tiny$\pm$0.0} \\
        FWM-AE & {2.33}{\color{lightgrey}\tiny$\pm$0.48} & {49.2}{\color{lightgrey}\tiny$\pm$13.4} & {3.16}{\color{lightgrey}\tiny$\pm$1.04} & {38.0}{\color{lightgrey}\tiny$\pm$19.1} \\
        \textbf{FWM (our)} & {\bf 1.27}{\color{lightgrey}\tiny$\pm$0.33} & {\bf 99.4}{\color{lightgrey}\tiny$\pm$0.3} & {\bf1.05}{\color{lightgrey}\tiny$\pm$0.18} & {\bf 99.4}{\color{lightgrey}\tiny$\pm$0.2} \\
        \midrule
        - No Edge Actions & {1.44}{\color{lightgrey}\tiny$\pm$0.52} & {98.0}{\color{lightgrey}\tiny$\pm$0.4} & {1.47}{\color{lightgrey}\tiny$\pm$0.34} & {98.9}{\color{lightgrey}\tiny$\pm$0.7} \\
        - 1 GNN Layer & {1.78}{\color{lightgrey}\tiny$\pm$0.17} & {95.5}{\color{lightgrey}\tiny$\pm$1.8} & {1.80}{\color{lightgrey}\tiny$\pm$0.12} & {96.6}{\color{lightgrey}\tiny$\pm$1.1} \\
        - No RGB & {\bf1.27}{\color{lightgrey}\tiny$\pm$0.43} & {96.5}{\color{lightgrey}\tiny$\pm$1.4} & {1.20}{\color{lightgrey}\tiny$\pm$0.62} & {98.8}{\color{lightgrey}\tiny$\pm$0.4} \\
        - No Coordinates & {11.64}{\color{lightgrey}\tiny$\pm$1.08} & {0}{\color{lightgrey}\tiny$\pm$0.0} & {9.82}{\color{lightgrey}\tiny$\pm$4.11} & {0}{\color{lightgrey}\tiny$\pm$0.0} \\
        - No Factorization & {10.19}{\color{lightgrey}\tiny$\pm$1.14} & {0}{\color{lightgrey}\tiny$\pm$0.0} & {8.39}{\color{lightgrey}\tiny$\pm$1.10} & {0}{\color{lightgrey}\tiny$\pm$0.0} \\
        \toprule
        \textbf{Method} & \multicolumn{2}{c}{\textbf{\textsc{Shapes}--Training}} & \multicolumn{2}{c}{\textbf{\textsc{Shapes}--Zero-Shot}} \\
         & \begin{tabular}{@{}c@{}} RMSE \\ (cm)\end{tabular} & \begin{tabular}{@{}c@{}} Hits@1 \\ (\%)\end{tabular} & \begin{tabular}{@{}c@{}} RMSE \\ (cm)\end{tabular} & \begin{tabular}{@{}c@{}} Hits@1 \\ (\%)\end{tabular} \\
        \midrule
        FWM-AE & {0.52}{\color{lightgrey}\tiny$\pm$0.32} & {86.6}{\color{lightgrey}\tiny$\pm$19} & {0.65}{\color{lightgrey}\tiny$\pm$0.37} & {82.9}{\color{lightgrey}\tiny$\pm$21} \\
        \textbf{FWM (our)} & {0.51}{\color{lightgrey}\tiny$\pm$0.10} & {\bf93.2}{\color{lightgrey}\tiny$\pm$5.4} & {0.68}{\color{lightgrey}\tiny$\pm$0.11} & {\bf93.1}{\color{lightgrey}\tiny$\pm$5.1} \\
        \midrule
        - No Edge Actions & {0.66}{\color{lightgrey}\tiny$\pm$0.09} & {80.9}{\color{lightgrey}\tiny$\pm$7.0} & {0.83}{\color{lightgrey}\tiny$\pm$0.08} & {83.0}{\color{lightgrey}\tiny$\pm$3.6} \\
        - 1 GNN Layer & {1.05}{\color{lightgrey}\tiny$\pm$0.07} & {72.0}{\color{lightgrey}\tiny$\pm$7.2} & {1.28}{\color{lightgrey}\tiny$\pm$0.10} & {73.6}{\color{lightgrey}\tiny$\pm$5.1} \\
        - No RGB & {\bf0.50}{\color{lightgrey}\tiny$\pm$0.27} & {85.8}{\color{lightgrey}\tiny$\pm$6.0} & {\bf0.64}{\color{lightgrey}\tiny$\pm$0.27} & {87.9}{\color{lightgrey}\tiny$\pm$3.6} \\
        - No Coordinates & {9.56}{\color{lightgrey}\tiny$\pm$0.97} & {0}{\color{lightgrey}\tiny$\pm$0.0} & {9.81}{\color{lightgrey}\tiny$\pm$0.95} & {0}{\color{lightgrey}\tiny$\pm$0.0} \\
        - No Factorization & {6.19}{\color{lightgrey}\tiny$\pm$0.81} & {0}{\color{lightgrey}\tiny$\pm$0.0} & {6.79}{\color{lightgrey}\tiny$\pm$0.82} & {0}{\color{lightgrey}\tiny$\pm$0.0} \\
        \bottomrule
    \end{tabular}
    \caption{Comparison between factored world models, baselines and ablations in \textsc{Cubes} and \textsc{Shapes}. We evaluate the models both on the tasks they were trained on (second and third column) and on unseen tasks without additional fine-tuning (fourth and fifth column). We report block position error (RMSE, the lower the better) and action sequence ranking score (Hits@1, the higher the better) (Section \ref{sec:experiments:training}). Each model was run with 4 random seeds and we report means and 95\% confidence intervals.}
    \label{tab:main_comp}
\end{table}

%with a heuristic that encodes elementary information about the nature of pick-and-place tasks.

\textbf{Pick-and-place heuristic for interchangeable objects.} We outline the heuristic in Algorithm \ref{alg:heuristic}. We use a trained latent transition model to predict the effect of action $a$ on encoded state $z$ (line 1). The individual objects in the predicted latent state $\hat{z}$ are then matched to objects in the goal state $z^g$ by the Hungarian matching algorithm \cite{kuhn55hungarian} (line 3). If the predicted state and the goal contain a different number of objects, the extracted objects that were not matched are discarded. For each pair of matched objects, the heuristic distance is assigned one of three values: (a) if the predicted object state is close enough to the goal state (controlled by a hyper-parameter $\delta$) then their distance is added to the heuristic (line 7); (b) if the object is in the robot's hand, it is at least one time step away from the goal state (line 9); (c) if the object is on the ground and far from the goal state, it is at least two time steps away (pick and place, line 11). We train an additional classifier $j_{\psi}$ to make binary predictions about an object being on-ground / in-hand based on the latent space of FWMs.

% heuristic for different objects
% assume to know the sequence of objects
\textbf{Pick-and-place heuristic for sequential tasks.} The above heuristic fails when a particular sequence of objects is required. For example, an action that picks up a roof of a structure that has not been built is assumed to be a distance of 1 away from the goal state of the roof, due to the heuristic being optimistic. For such cases, we add the assumption that we know the order in which to place the objects. Hence, the role of the model is to plan the robot actions in order to execute on the sequence. Since we factor a state into objects, the sequence in which the objects are manipulated can easily be extracted from a single demonstration of a particular task. In this case, Algorithm \ref{alg:heuristic} no longer matches objects between the current state and the goal (line 3). An object is a distance of one to its goal (line 9) only if the previous objects in the sequence were correctly placed. Please see Algorithm \ref{alg:heuristic2} (Appendix).

% how we use an ensemble
Since our search procedure finds the best possible outcome, it can target incorrect predictions of the transition model. This problem can be mitigated by using an ensemble of models. To evaluate the heuristic value of an action in an ensemble, we apply Algorithm \ref{alg:heuristic} to each member of the ensemble separately. Then, we take the average over all member heuristic. Members in the ensemble only differ in the random seed used to initialize their weights before training. The ensemble could be further improved by increasing its diversity.

\begin{table}
    \centering
    \begin{tabular}{llrrrr}
        \toprule
         & & \multicolumn{2}{c}{\textbf{\textsc{Cubes}, Success (\%)}} & \multicolumn{2}{c}{\textbf{\textsc{Shapes}, Success (\%)}} \\
        \textbf{Agent} & \textbf{\# Models} & \textbf{Train} & \textbf{Zero-Shot} & \textbf{Train} & \textbf{Zero-Shot} \\
        \midrule
        FWM-H & 20 & {\bf95.0}{\color{lightgrey}\tiny$\pm$0.0} & 75.0{\color{lightgrey}\tiny$\pm$22.6} & {\bf70.0}{\color{lightgrey}\tiny$\pm$9.4} & {\bf67.5}{\color{lightgrey}\tiny$\pm$21.5} \\
        FWM-H & 10 & 90.0{\color{lightgrey}\tiny$\pm$0.0} & {\bf80.0}{\color{lightgrey}\tiny$\pm$7.9} & 67.5{\color{lightgrey}\tiny$\pm$9.0} & 62.5{\color{lightgrey}\tiny$\pm$19.7} \\
        FWM-H & 1 & 35.0{\color{lightgrey}\tiny$\pm$0.0} & 23.8{\color{lightgrey}\tiny$\pm$16.3} & 23.1{\color{lightgrey}\tiny$\pm$16.2} & 17.5{\color{lightgrey}\tiny$\pm$17.1} \\
        FWM-$L^2$ & 20 & 30.0{\color{lightgrey}\tiny$\pm$0.0} & 1.2{\color{lightgrey}\tiny$\pm$2.2} & 27.5{\color{lightgrey}\tiny$\pm$26.7} & 25.0{\color{lightgrey}\tiny$\pm$22.5} \\
        \bottomrule
    \end{tabular}
    \caption{Aggregated results for heuristic search with an ensemble of factored world models. The results are averages over training and zero-shot transfer tasks. Table \ref{tab:tasks} shows the goal states of the tasks as well as a breakdown of FWM-H with 20 models. FWM-H refers to using our heuristic from Section \ref{sec:methods:search}; FWM-$L^2$ is a baseline heuristic that computes the $L_2$ distance between the encoded current state and the encoded goal state.}
    \label{tab:planning}
\end{table}

\section{Experiments}
\label{sec:experiments}

\begin{figure}[t]
  \centering
  \includegraphics[width=0.8\linewidth]{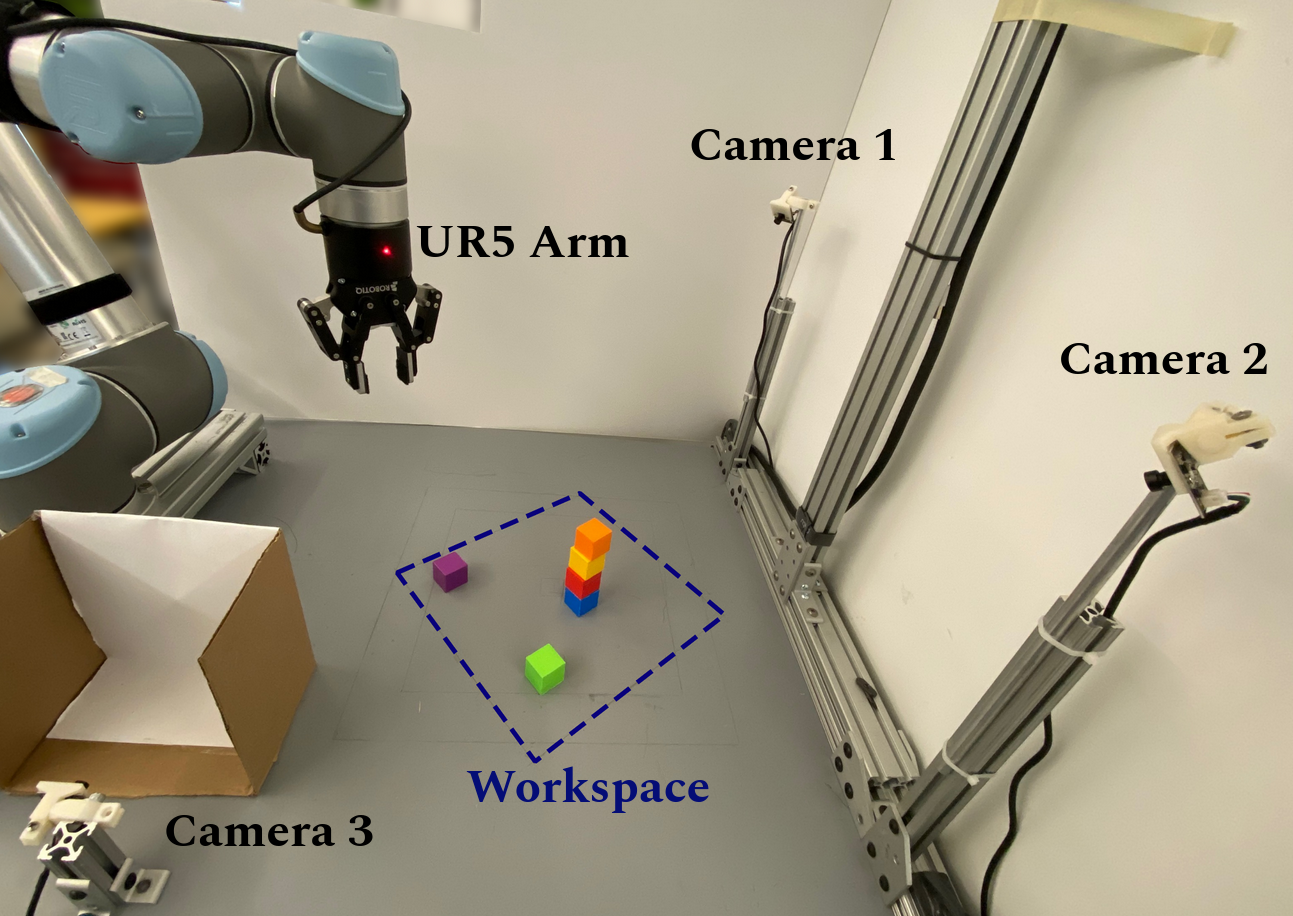}
  \caption{Our real-world setup includes an UR5 arm, two side-viewing cameras (Camera 1 and 2), a camera that takes an image of the contents of the robot's gripper (Camera 3) and a workspace with six cubes.}
  \label{fig:real_env}
\end{figure}

Our empirical evaluation focuses on pick-and-place robotic manipulation tasks performed in simulation as well as on a UR5 robotic arm (Figure \ref{fig:envs} and Figure \ref{fig:real_env}). The environments involve a robotic arm manipulating objects of four shapes (cube, brick, triangle, roof) in a $30{\times}30$ cm workspace. The agent controlling the arm uses pre-defined \texttt{pick} and \texttt{place} motion primitives: it chooses a particular $(x, y)$ location (continuous action space) in the workspace in which the arm executes a top-down pick or place action depending on if it is holding an object.

The state of the environment is captured by two RGB cameras pointed at the workspace and a third RGB camera that captures the content of the robotic gripper. We assume access to a bounding box for each object: we derive it from the ground-truth state of the simulator if available; we use color segmentation with uniquely colored blocks in the real world. The factored state $s = \langle s_1, s_2, ..., s_K \rangle$ is created by cropping the contents of each bounding box with added padding and resizing the resulting image to an $18{\times}18$ square. We add bounding box coordinates into the cropped image in the form of four additional coordinate grids channels (Figure \ref{fig:observation}, Appendix \ref{sec:app:env_details}). Hence, each view of each object results in an $18{\times}18$ image with 3 RGB and 4 coordinate grid channels. Note that we only require object bounding boxes within the images taken by side-viewing cameras. We do not need the ground-truth $(x,y,z)$ object positions in order to generate the factored state. We concatenate the two views of each object channel-wise, creating a 14-channel image for each object. If an object is held by the robotic gripper, we replace the two views with two images of the gripper --- one from the front and one from the side. These two images can be captured by a single camera by rotating the robotic arm in between images. The coordinate grids for hand images are set to zero.

\begin{table}[t]
    \centering
    \begin{tabular}{lccccc}
        \toprule
         & \multicolumn{5}{c}{\textbf{Real World Tasks}} \\
        \midrule
         & \includegraphics[width=0.05\textwidth]{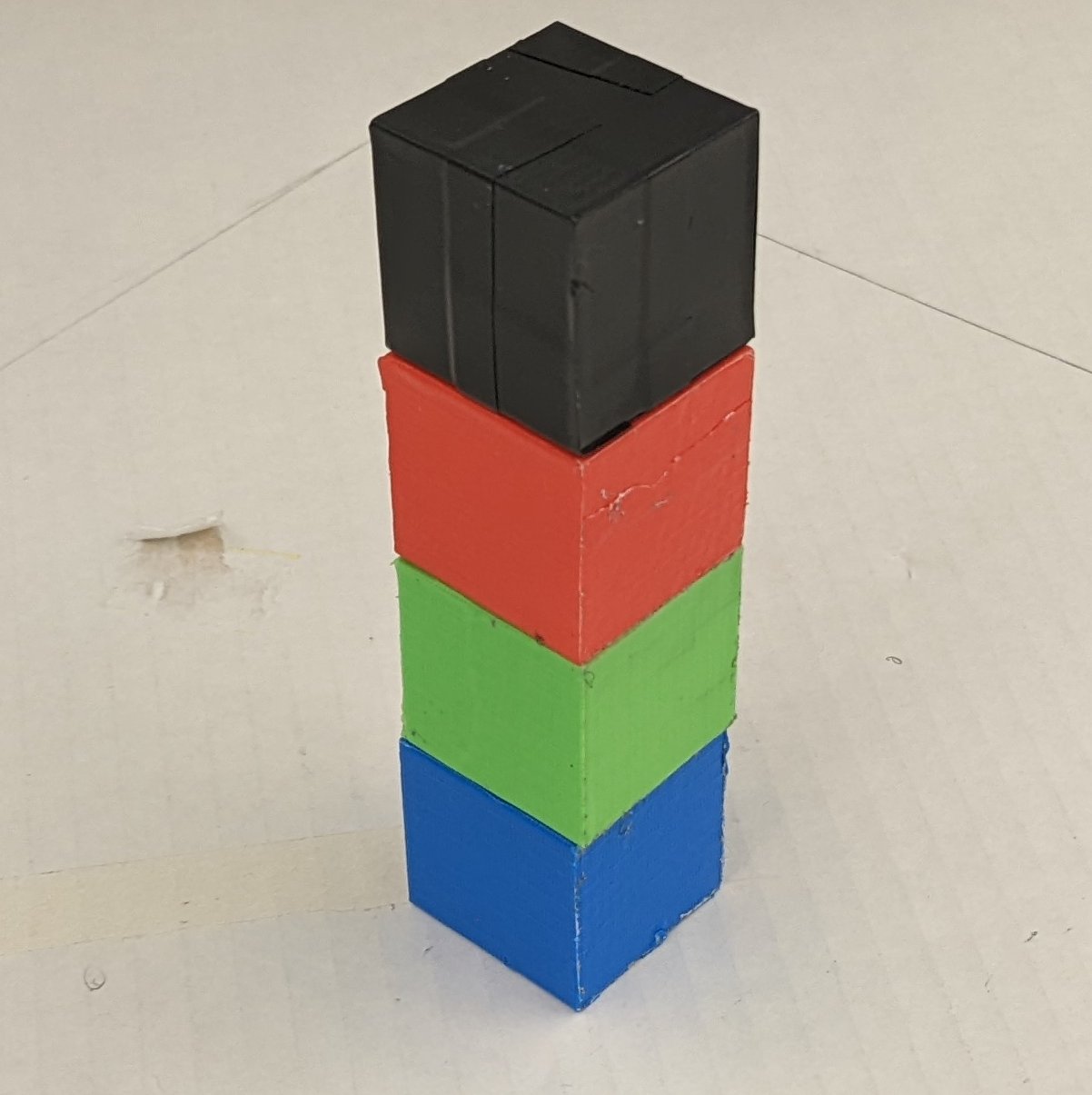} & \includegraphics[width=0.05\textwidth]{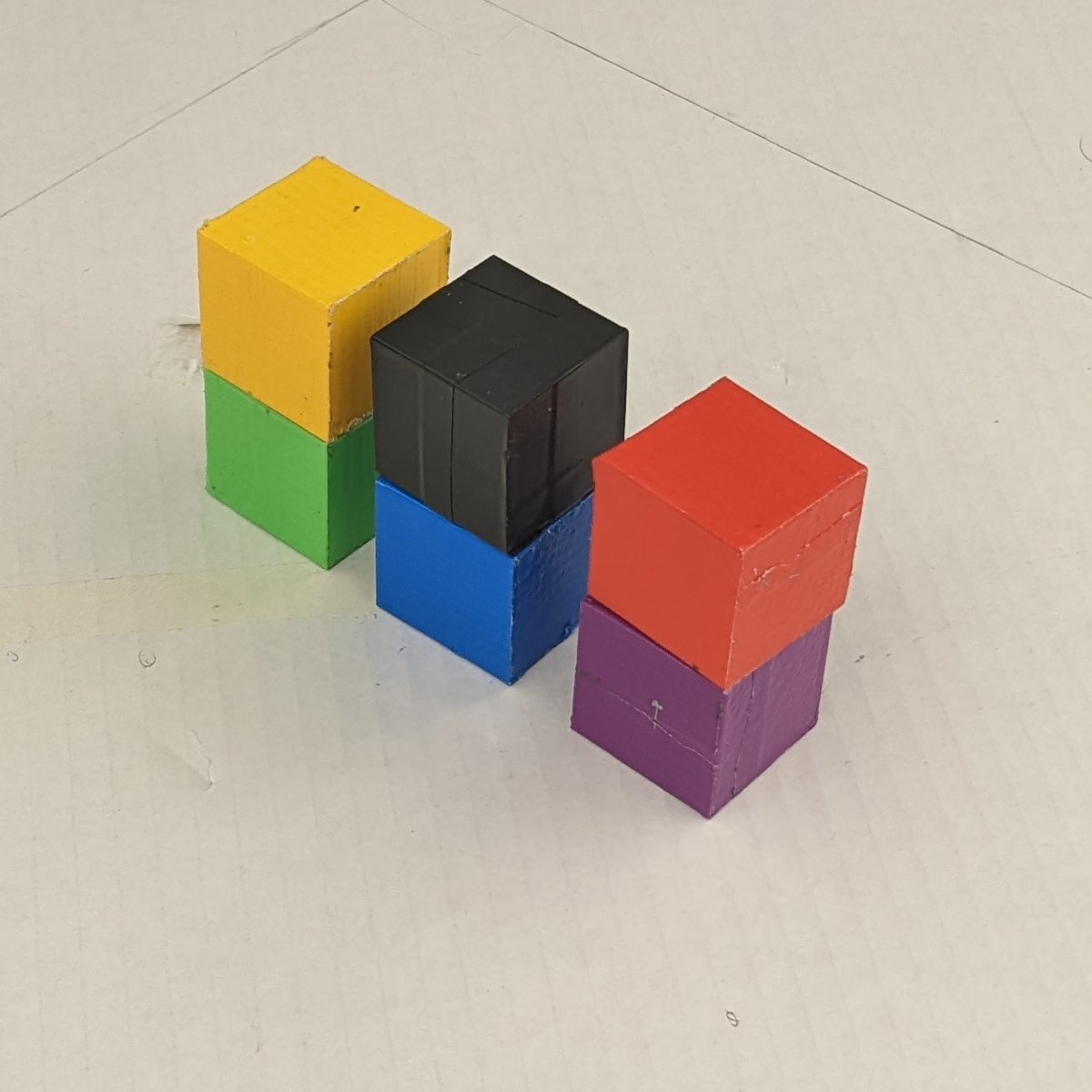} & \includegraphics[width=0.05\textwidth]{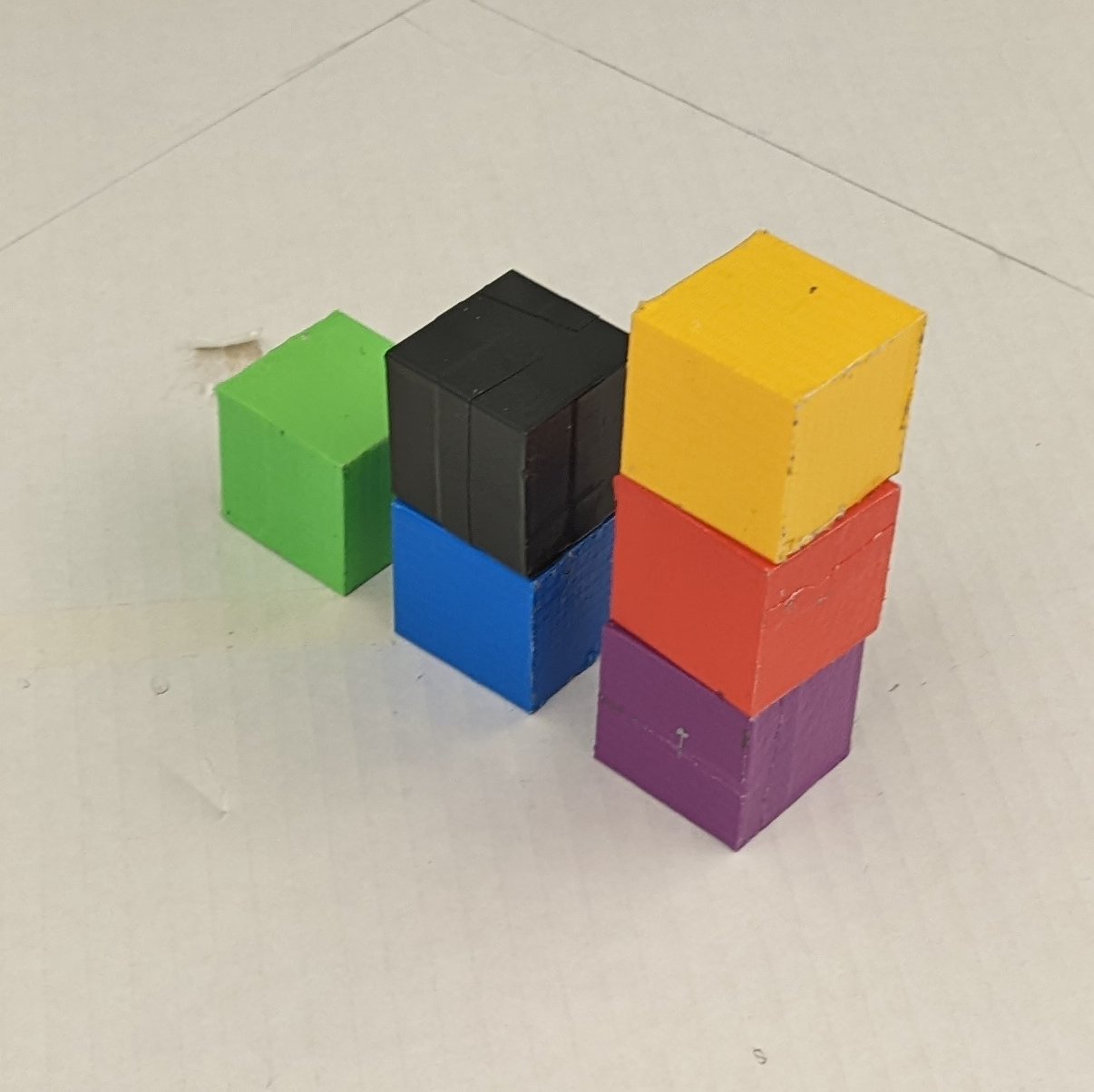} & \includegraphics[width=0.05\textwidth]{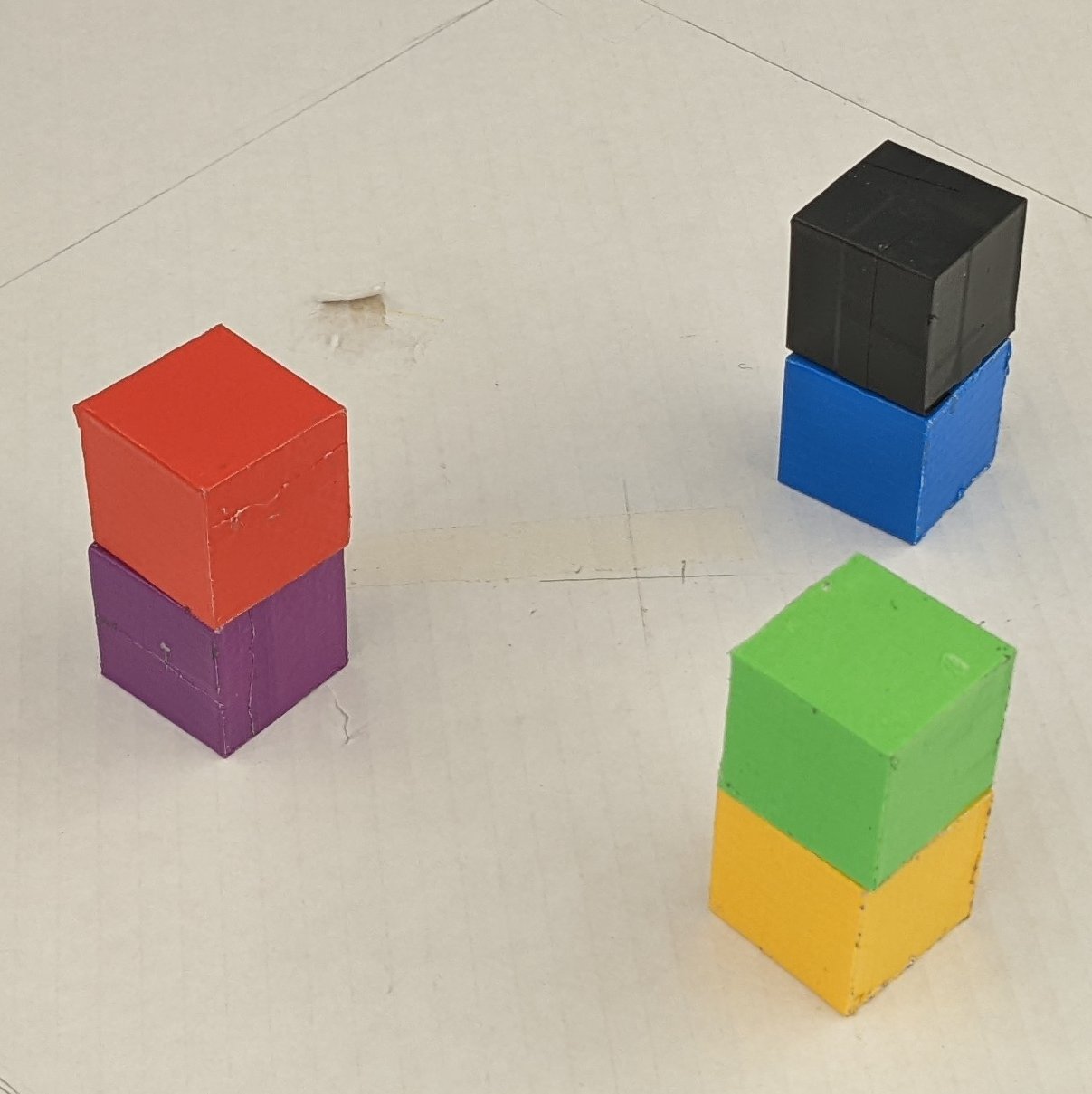} & \includegraphics[width=0.05\textwidth]{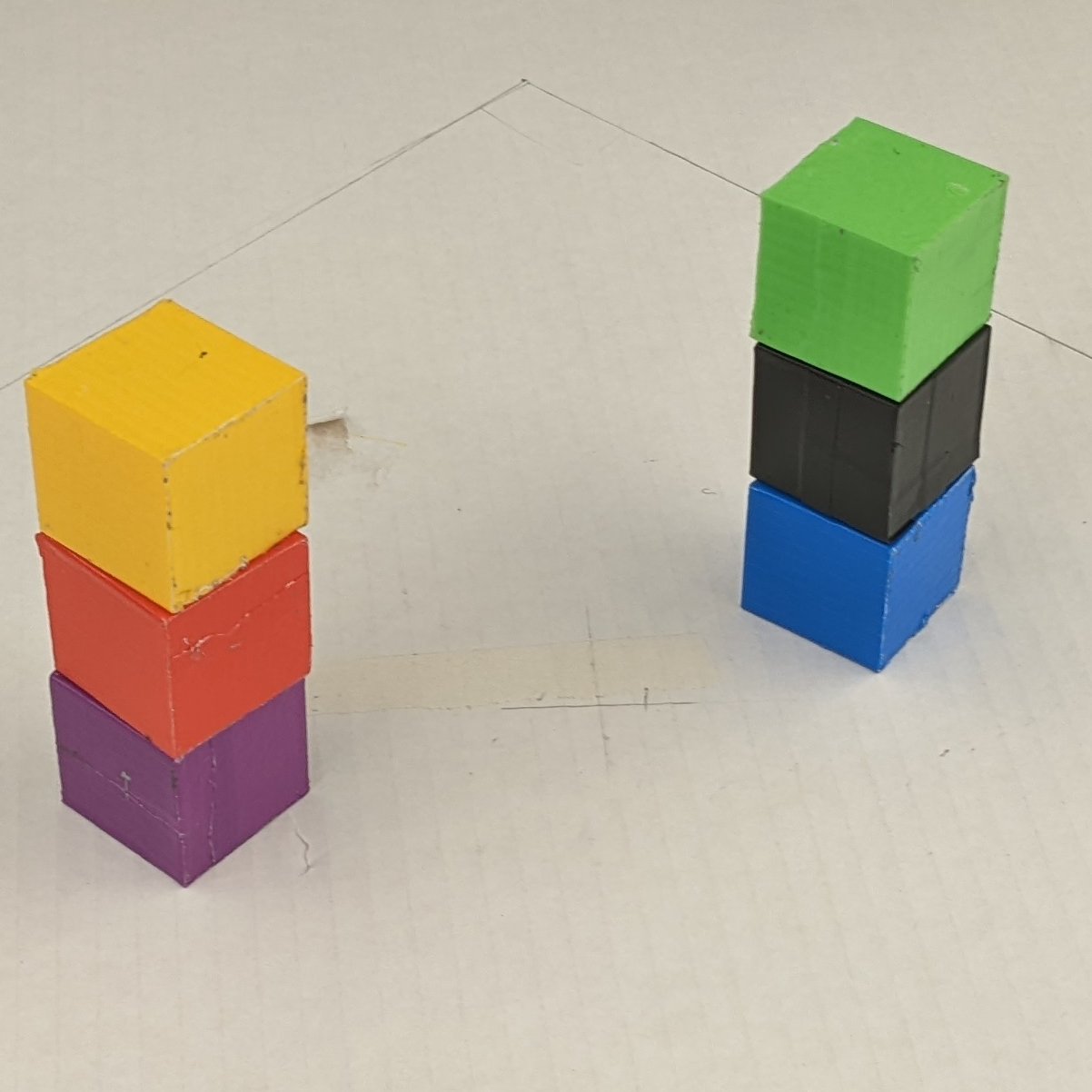} \\
        Success & 2/6 & 2/6 & 1/6 & 3/6 & 4/6 \\
        Progress & 79\% & 89\% & 83\% & 89\% & 92\% \\
        \bottomrule
    \end{tabular}
    \caption{Success rates of an ensemble of 20 FWMs trained in simulation and deployed on a physical robot. Progress (\%) refers to the percentage of blocks that were successfully placed. The agent often fails when placing the last cube in the goal structure.}
    \label{tab:planning_real}
\end{table}

The environment poses a difficult exploration problem: a sequence of random actions is unlikely to create a large structure without knocking it over. A model-free agent with a random exploration policy fails to learn to build structures involving more than two objects (\cite{biza21action}, Table 2, "DQN RS"). Therefore, we collect a dataset of demonstrations from an experiment with added randomness. We evaluate both the ability of FWMs to fit tasks with expert trajectories as well as the ability to generalize to a set of held-out tasks \textit{without fine-tuning} (zero-shot transfer, Table \ref{tab:tasks}). We further demonstrate that FWMs can be used to plan for novel tasks.

We aim to answer the following questions:
\begin{itemize}
    \item Can FWMs learn the physics of robotic pick-and-place? (Section \ref{sec:experiments:training})
    \item Does permutation equivariance in FWMs facilitate generalization to unseen tasks? (Section \ref{sec:experiments:novel_tasks})
    \item What is the contribution of the individual components of FWMs to their performance? Which component of the factored state is the most important?  (Section \ref{sec:experiments:ablations})
    \item Can FWMs be used for planning? Can FWMs trained in simulation transfer to a physical robot? (Section \ref{sec:experiments:planning}).
\end{itemize}

\subsection{Learning Robotic Manipulation Dynamics}
\label{sec:experiments:training}

\textbf{Setup.} To train FWMs, we collect datasets from the training tasks in \textsc{Cubes} and \textsc{Shapes} (Table \ref{tab:tasks}). Each dataset consists of 200k transitions collected by an expert with added randomness, see Appendix \ref{sec:app:experts} for details. FWMs learn for 200 epochs using contrastive learning, and are subsequently evaluated both on the training and zero-shot transfer tasks. We use a block position prediction metric to evaluate the quality of the learned representation as well as the latent transition model. We also evaluate the model in a setting where it predicts the outcomes of several action sequences. The action sequences are similar, but only one of them reaches the goal state of a given task. By predicting the outcome of each sequence, the model guesses which one it is. We use this setting as a proxy for planning.

\underline{Block position prediction (RMSE):} After training, we freeze the model and train an additional decoder to predict the $(x, y, z)$ position of the center of each object from its latent representation. The decoder, a
Multi-Layer Perceptron with two hidden layers, is trained for five epochs. It is trained using additional supervision (pairs of states and position labels) not available during training of factored world models. We evaluate both the model's ability to represent block positions in the current state ($t=0$) and its ability to make predictions for trajectories ($t>0$). We report the root mean squared error in centimeters. The quantity predicted in this setting is different than the bounding box coordinates provided to the model as input. The bounding box coordinates index into a flat 2D images; the model needs to account for perspective to predict the \texttt{(x,y,z)} positions of objects.

\underline{Action sequence ranking (Hits@1, MRR):} We start with an optimal action sequence that achieves a goal state of a particular task. Then, we generate ten other action sequences where each action is perturbed by noise with magnitude $\epsilon$. The noise is added to each action in the sequence by drawing $\epsilon \sim \text{Unif}[0, \epsilon]$ and $\theta \sim \text{Unif}[0, 2\pi]$ that are then combined to create a 2D vector in the direction of $\theta$ with a length of $\epsilon$ that is added to the \texttt{(x,y)} coordinate of each action. Each perturbed action sequence must not achieve the given task; otherwise, we re-sample noise. The model observes the starting state and predicts the final state of each action sequence. We also encode the final state of the correct action sequence and compare its distance to the \textit{predicted} final states of all action sequences. Hits@1 report the fraction of times the model's prediction of the final state of the correct action sequences was closer to the encoded final state than the predictions for all of the incorrect action sequences. Since the incorrect action sequences are chosen to be a small distance from the correct one, a model making random predictions is not expected to reach Hits@1 of 1/11. In fact, the model needs to be fairly competent to get a non-zero score.

\textbf{Results.} We report results for training tasks in Table \ref{tab:main_comp} top section and Figure \ref{fig:action_eps} left. FWM reaches low block prediction error (1.27 cm for \textsc{Cubes} and 0.51 cm for \textsc{Shapes}; the size of a cube is 3 cm for comparison) and high action ranking score (99\% for \textsc{Cubes} and 93\% for \textsc{Shapes}). We compare to an autoencoder baseline (FWM-AE) and the Region Proposal Interaction Network (RPIN, \citet{qi21learning}) with minor changes. For FWM-AE, we add a decoder to our model and train with an autoencoding loss (specifically the Embed to Control loss by  \citet{watter15embed}) instead of contrastive learning. The Embed to Control loss consists of three terms: reconstruction of the current and predicted latent state, and a $L^2$ between the predicted and next encoded state. To adapt RPIN, a factored video prediction method, we append an \texttt{(x,y,pick/place)} action to the input of the Prediction module in their Convolutional Interaction Network. We recreate our dataset in the same format as their Shape Stacking experiment ($224{\times}224$ images with a bounding box for each object) and we use the same hyper-parameter.

Compared to an autoencoding loss (FWM-AE), contrastive learning (FWM) succeeds both in \textsc{Cubes} and \textsc{Shapes}, whereas the autoencoding loss fails to fit \textsc{Cubes}. In \textsc{Shapes}, autoencoding is on par with contrastive loss in terms of block position prediction, but underperforms in action sequence ranking. We believe the difference can be explained by the focus of the two different losses: autoencoding loss directly incentivizes the model to reconstruct the state of the environment (which include bounding box coordinates in the form of a coordinate grid), whereas contrastive loss focuses on learning compact state representations predictive of the future. We plot action ranking Hits@1 as a function of noise level $\epsilon$ for \textsc{Shapes} in Figure \ref{fig:action_eps} left. Contrastive loss outperforms autoencoding for all noise levels with the area under the curve being 97\% for FWM and 44\% for FWM-AE. 

Our adaptation of RPIN to action-conditioned sequence prediction fails to learn a dynamics model. The encoder of RPIN can capture some position information: it reaches around 6 cm RMSE for the task of predicting ground-truth object coordinates without forward modeling. However, the forward model fails, reaching 20 cm RMSE. Our results for C-SWM are discussed in Appendix \ref{sec:app:additional}.

\subsection{Zero-Shot Generalizing to Novel Tasks}
\label{sec:experiments:novel_tasks}

\textbf{Setup.} We further evaluate models trained in Section \ref{sec:experiments:novel_tasks} on tasks unseen during training (Table \ref{tab:tasks}, right). We specifically make sure that the training datasets do not contain a single example of a goal state of the novel tasks. In \textsc{Cubes}, the training task is building a 4-stack and the novel tasks are building a wall, stairs, a wall on ground and stairs on ground. We found that the training data collection policy never solves the novel tasks (we use a separate policy to collect evaluation trajectories for testing tasks). In \textsc{Shapes}, we train on half of the possible structures with a height of three and roof on top, and test generalization on the other half. In this case, the data collection policy for the eight training tasks solves one of the testing tasks around once every 50 episodes (due to added randomness) and we delete these episodes. We report block position error and action ranking Hits@1 described in Section \ref{sec:experiments:training}. We do not perform any fine-tuning on the novel tasks in this experiment.

\textbf{Result.} In both \textsc{Cubes} and \textsc{Shapes}, FWM transfers to the novel tasks with only a marginal decrease (and in some cases an increase) in performance. We consider this to be an important demonstration of the generalization properties of permutation equivariant models. In contrast, only limited generalization and transfer properties have been shown in prior work \citep{li20towards,biza21action}. FWM-AE suffers a large decrease in action sequence ranking while outperforming FWM on block position prediction in \textsc{Shapes}. We plot action ranking error as a function of $\epsilon$ for \textsc{Shapes} in Figure \ref{fig:action_eps} right; we see a pattern analogous to results for training tasks (Section \ref{sec:experiments:training}). The areas under curve are 96\% for FWM and 28\% for FWM-AE.

\begin{table}[t!]
    %\centering
    \begin{tabular}{llll}
        \toprule
        Pick-and-place tasks & \textbf{FWM (Our)} & OP3 \citep{veerapaneni19entity} & ReNN \citep{li20towards} \\
        \midrule
        Number of objects modeled & \textbf{8} & 3 & 10 \\
        Maximum time horizon (*) & \textbf{12} & 2 & 12 \\
        Works with image states & \textbf{\textcolor{yes_color}{Yes}} & \textcolor{yes_color}{Yes} & \textcolor{red}{No} \\
        Learns a transition model & \textbf{\textcolor{yes_color}{Yes}} & \textcolor{yes_color}{Yes} & \textcolor{red}{No} \\
        Learns to factor states & \textbf{\textcolor{red}{No}} & \textcolor{yes_color}{Yes} & \textcolor{red}{No} \\
        Shows zero-shot transfer & \textbf{\textcolor{yes_color}{Yes}} & \textcolor{red}{No} & \textcolor{yes_color}{Yes} \\
        Controls a physical robot & \textbf{\textcolor{yes_color}{Yes}} & \textcolor{red}{No} & \textcolor{red}{No} \\
        \bottomrule
    \end{tabular}
    \caption{Comparison between our method and two recent works on object-factored planning or policy learning. (*) here we only consider pick-and-place tasks that can be solved with at least 70\% success rate.}
    \label{tab:planning_comparison}
\end{table}

\subsection{FWM Ablations}
\label{sec:experiments:ablations}

We report results for the following ablations in Table \ref{tab:main_comp}: using only one graph neural network layer (1 GNN Layer), not providing actions to the edge networks in each GNN (No Edge Actions), only using bounding box coordinates as inputs (No RGB), only using RGB images as inputs (No Coordinates) and using a monolithic latent transition model (No Factorization).

An interesting result in our ablation studies is the comparison of FWMs with and without RGB images. The No RGB baseline has access only to the coordinates of the bounding boxes for each object. Nevertheless, it is able to make predictions about the positions of objects (RMSE) on par with the full model. But, there is a significant decrease in its ability to rank action sequences (Hits@1). We believe more complex object interaction (e.g. packing a bin with household objects, building from LEGO) would put a higher emphasis on images. For \textsc{Cubes} and \textsc{Shapes}, the No RGB version of FWMs is a valid option when the distribution of images changes between training and evaluation--we use it in our real-world robot planning experiments.

Both the 1 GNN Layer and No Edge Actions ablations confirm that the architecture of FWMs does not have redundant elements. No Factorization and No Coordinates ablations show that both of these components are vital to FWMs. Without coordinate grids, the model perceives images of objects without any information about where they are located in the environment. Without object factorization, the monolithic latent transition model is unable to generalize beyond the specific arrangements of blocks seen in the training tasks. Here, we mean in-distribution generalization to building known structures in novel locations of the workspace. Naturally, the monolithic model also fails in zero-shot transfer.

\subsection{Planning with FWMs}
\label{sec:experiments:planning}

\textbf{Setup.} We use an ensemble of 20 FWMs and the pick-and-place heuristic described in Section \ref{sec:methods:search}. At each step, the agent evaluates 10k actions arranged in a grid spanning the \texttt{pick(x,y)/place(x,y)} action space. The pick/place action primitive is chosen based on the state of the robot's gripper. The agent receives a single example of a particular goal state and has 20 time steps in simulation and 30 in the real world to build it. Planning is done online at each time step (i.e. closed-loop control). For our real-world experiment, we do not fine-tune on real-world data. Instead, we use a version of FWMs that operates on bounding boxes only (the No RGB ablation), which allows us to avoid the problem of generalizing to real-world images. The height at which a pick or place action is performed is based on top-down depth images.
%Table \ref{tab:planning_comparison} compares the capabilities of our agent to recent works on object-factored planning and model-free learning \citep{veerapaneni19entity,li20towards}.

\textbf{Result.} Table \ref{tab:tasks} breaks down the planning success rates for an ensemble of 20 FWMs; results are aggregated in Table \ref{tab:planning}. Using an ensemble is essential to an accurate planner: an ensemble of 10 models more than triples the success rate of a single model. Considering we evaluate 10k actions to make a single decision, it is unsurprising that the planner often finds a gap where the model predicting an unreasonably good outcome. In contrast, difficult-to-predict regions of the action space (such as situation where the gripper hits an object instead of picking it up) tend to have a high variance in terms of the heuristic distances predicted by each member of an ensemble. Hence, the ensemble chooses safer and more predictable actions. Further, we find that using a pick-and-place heuristic is crucial to successful planning. FWM-$L^2$ (Appendix Algorithm \ref{alg:heuristic_l2}), a baseline that computes the heuristic distance as an $L^2$ distance between the current encoded state and the encoded goal state, fails to solve the majority of zero-shot transfer tasks. It tends to push objects around instead of picking and placing them on top of each other.

We further discuss the relation of FWMs to recent object-factored agents in Table \ref{tab:planning_comparison}. We attempted to adapt ReNN \citep{li20towards} to our spatial action space, but it failed to solve tasks beyond placing two cubes on top of each other. ReNN includes design decisions appropriate to high-frequency control of robot motions, such as a reward shaped to encourage the robot to pick-and-place objects (it has a similar aim to our pick-and-place heuristic, but different execution). These decisions are not trivially applicable to a pick-and-place actions space. Moreover, we use on the order of 10 to 100 times fewer environment steps (depending on how many environment steps we consider each pick and place action to consist of).

\begin{figure}[t!]
    \centering
    \includegraphics[width=\linewidth]{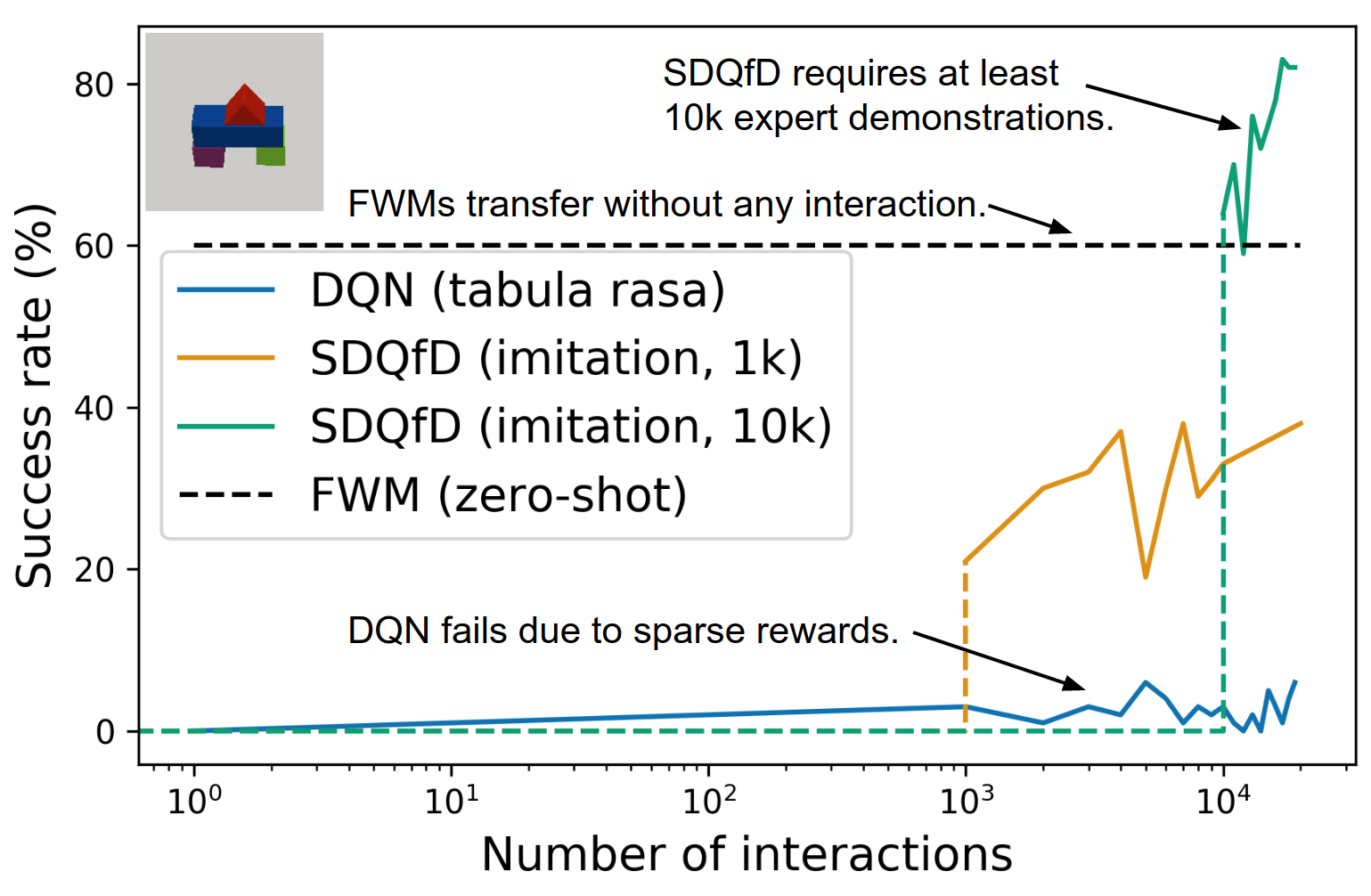}
    \caption{Success rates of an ensemble of 20 FWMs, a vanilla deep Q-network \citep{mnih15human} and SDQfD \citep{wang20policy} in a transfer task from \textsc{Shapes} shown in the top-left corner. FWMs have not seen this task before, but were trained on other tasks from \textsc{Shapes}. DQN starts with random weights and SDQfD receives 1k or 10k expert demonstration on the target task.}
    \label{fig:planning_comparison}
\end{figure}

Figure \ref{fig:planning_comparison} highlights the difference between our zero-shot transfer approach (using FWMs with heuristic search) and other model-free methods. We plot performance on a randomly chosen zero-shot transfer task from \textsc{Shapes} (FWMs never encountered it). A vanilla deep Q-network does not learn due to the sparsity of rewards (reward of 1 for reaching the goal, otherwise 0). SDQfD, an imitation learning method for spatial action spaces, eventually outperforms FWMs, but it requires a large number of expert demonstrations of each new task \citep{wang20policy}. The implementation and hyper-parameters of the DQN and SDQfD baselines are based on \citep{biza21action}.

Finally, we report performance on a real-world robot (Table \ref{tab:planning_real}, Figure \ref{fig:real_env}). An ensemble of 20 FWMs trained in simulation and deployed on the physical robot succeeds in 40\% of trials and builds 85\% of the goal structure on average. We find that our agent tends to make mistakes at the end of the task, usually due to a compounding error of imprecise picks and places. Conversely, the agent makes correct high-level decisions about which cubes to pick and where to place them. We believe domain randomization, both in terms of the image state and bounding boxes as well as the physics, would make the agent more robust in the real world.

%Figure \ref{fig:planning} reports average success rates for training and zero-shot transfer in Cubes and Shapes. In the former, the agent can build a 4-stack (the only training task) almost perfectly, but sometimes struggles to arrange blocks in the tight formation that is required to complete the zero-shot transfer tasks. Conversely, we see only a small drop in performance between the training and transfer tasks in Shapes, with an average success rate of around 67\%. \edit{We are still working on finishing up the real-world experiments. Moreover, I sketched a potential comparison with model-free methods in Figure \ref{fig:planning_comparison}.}

\section{Conclusion} The results in this paper demonstrate that factorized world models can successfully be applied to manipulation tasks with continuous state and action spaces. The resulting models are able to learn transition dynamics that generalize to previously unseen sequences of actions and configurations of objects at test time, and are able to predict outcomes at moderate time horizons of up to 12 actions. We further show that an ensemble of FWMs can be used to plan for previously unseen tasks with around 70\% success rate in simulation and 40\% success rate on a physical robot without fine-tuning on real-world data. This represents a significant step in applying these models in practical robotics tasks.

%% Use plainnat to work nicely with natbib. 

\bibliographystyle{plainnat}
\bibliography{paper}

\newpage
\onecolumn
\appendix
\section{Experiment Details}

\subsection{Environment Details}
\label{sec:app:env_details}

Our PyBullet simulation consists of a UR5 robotic arm with a Robotiq gripper operating over a $30{\times}30$ cm workspace. In comparison, the size of a cube is $3{\times}3$ cm. The environment is captured from two sides by two cameras. Each camera produces a $90{\times}90$ image. PyBullet also provides a segmentation mask for each object, which we use to create bounding boxes. We first draw the smallest rectangular box that captures the whole segmentation mask and add 4 px symmetric padding to it. If the bounding box is smaller than $18{\times}18$ px, then they are padded up to the minimum size. We crop the RGB image and four coordinate grids inside of the bounding box and resize them to an $18{\times}18$ image. The aspect ratio of the RGB image gets corrupted in this step, but the model can still capture the size of each object based on the coordinate grids. We have a vertical and a horizontal coordinate grid that traverses the interval of [-1, 1] either left-to-right or up-to-down. We also create two additional coordinate grids that traverse the interval right-to-left and down-to-up.

\begin{figure}
    \centering
    \includegraphics[width=1\textwidth]{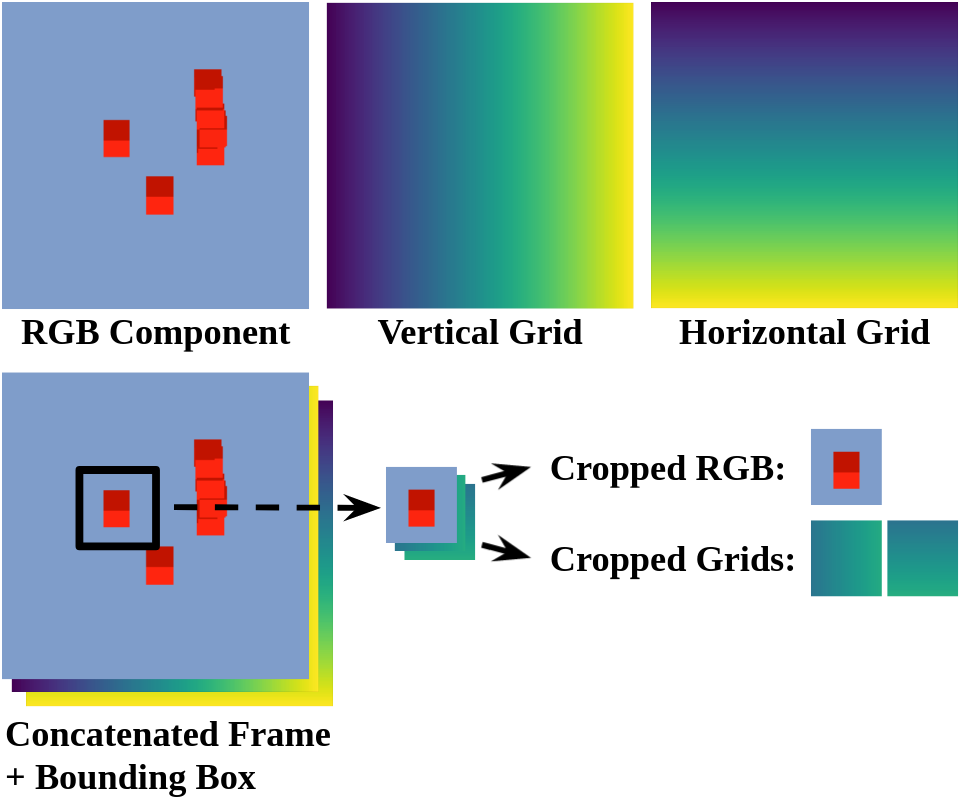}
    \caption{Schema of our pipeline for creating cropped object images. We concatenate an RGB image with horizontal and vertical coordinate grids. Using a bounding box with added padding, we crop an $18{\times}18$ image from both the RGB component and the coordinate grid components. By observing the coordinate grids, our agent known where \textit{in the image} the object was cropped. Note that the coordinate grids are derived from object bounding boxes, not the actual \texttt{(x,y,z)} object positions in the environment. Hence, we do not need to know the ground-truth object positions in order to generate the factored states. We add two additional coordinate grids by mirroring the vertical and horizontal grids (similar to positional encodings in \citet{locatello20object}).}
    \label{fig:observation}
\end{figure}

\begin{figure}[t!]
    \centering
    \begin{subfigure}{0.48\textwidth}
        \centering
        \includegraphics[width=1\textwidth]{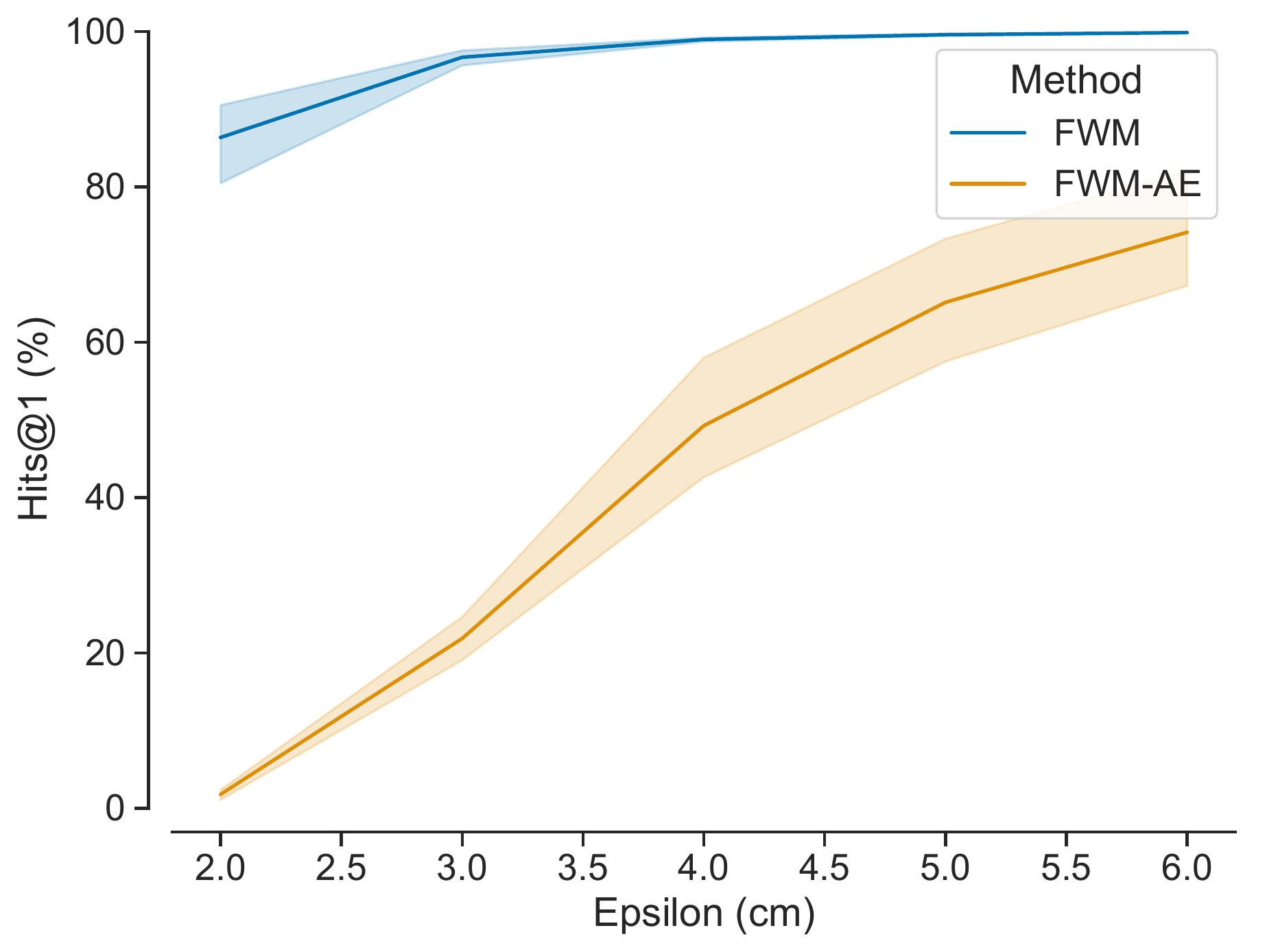}
    \end{subfigure}
    \begin{subfigure}{0.48\textwidth}
        \centering
        \includegraphics[width=1\textwidth]{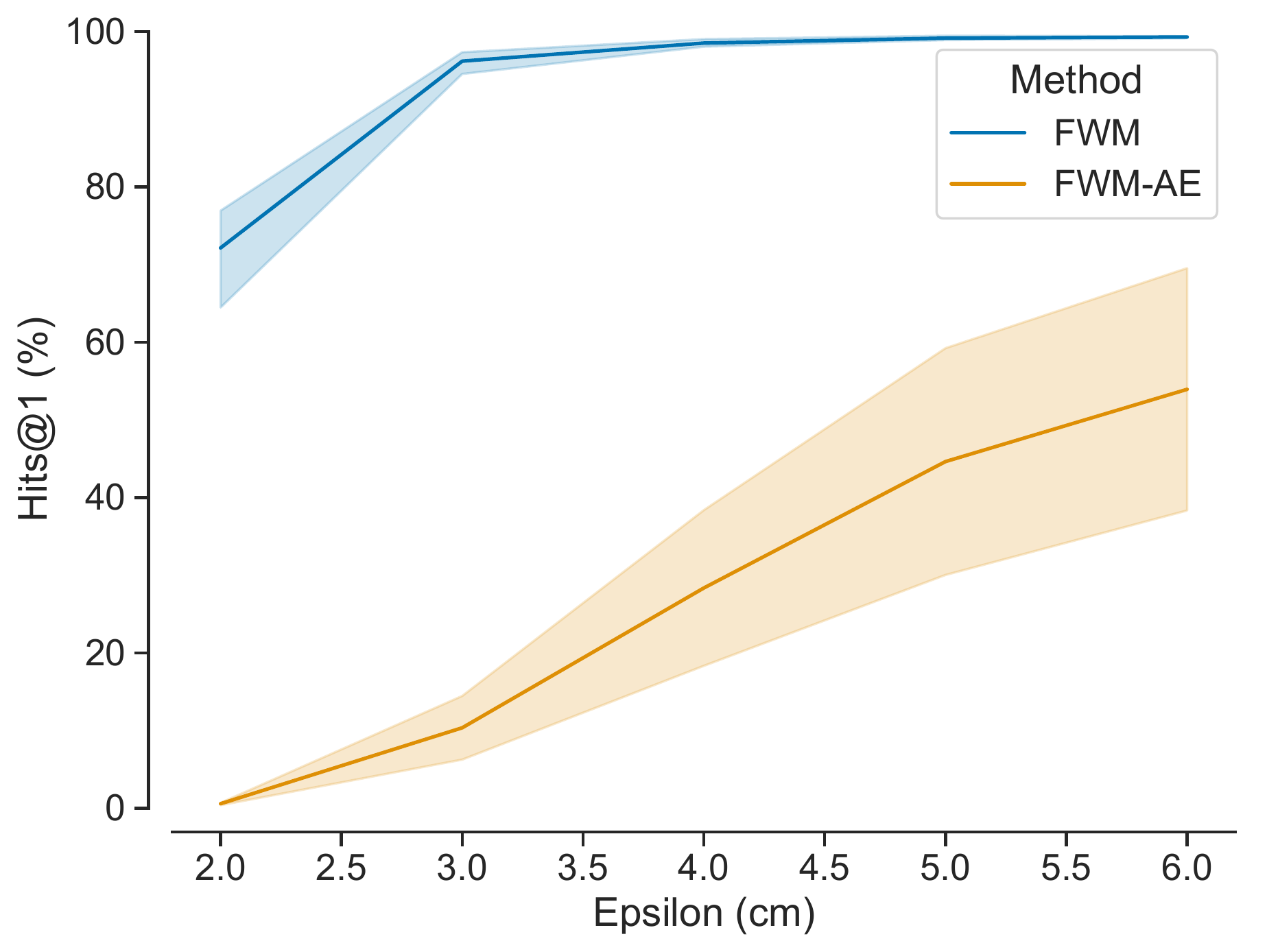}
    \end{subfigure}
    \caption{Action sequence ranking Hits@1 conditioned on action noise $\epsilon$. The noise level controls the degree to which negative action sequences differ from the position one (Section \ref{sec:experiments:training}). Left: results for training tasks; right: results for zero-shot generalization. We compare FWMs against baselines from Table \ref{tab:main_comp}. Means and 95\% confidence intervals over 5 random seeds are reported.}
    \label{fig:action_eps}
\end{figure}

\textbf{Block colors.} We use a single block color (red) for all experiments in simulation. In the real-world experiments, we assign each block a different color so that we can get bounding boxes using color segmentation (Figure \ref{fig:real_env}). When we train models in simulation that are supposed to transfer to the real world, we use a different color for each object that correspond to the colors used in the real-world. The decision between using same or different colors has a negligible impact on model performance, but we always assign a particular color to the some object slot.

\textbf{Physical robot.} We use a UR5 robot with a Robotiq 2F-85 gripper. The cubes are 3cm wide and are each covered with uniquely colored tape.  We use two ELP Megapixel USB Cameras with 100 degree lens, which are placed to view the center of the workspace from two sides.  Bounding boxes for each colored cube are extracted from the camera images by masking in HSV color space.

\subsection{Expert Policies}
\label{sec:app:experts}

\textbf{Experts for training datasets.} In \textsc{Cubes}, we have the following policy: with 70\% probability either pick a random cube that is not covered by other objects or place a cube on top of any other cube. Otherwise, execute a pick or place action in a random coordinate. Additionally, 1 cm random noise is added to the \texttt{(x,y)} position of each action. The agent does not have to decide between pick and place actions: hand empty $\rightarrow$ pick, hand full $\rightarrow$ place.

In \textsc{Shapes}, we use trained SDQfD agents provided by \citet{biza21action}. With 80\% probability we execute the actions predicted by the expert. Otherwise, during pick, we pick a random object with 50\% and we execute a pick at a random location otherwise; during place, we place in a random position.

We collect 200k training transitions for \textsc{Cubes} and \textsc{Shapes}.

\textbf{Experts for generalization datasets.} We simply collect optimal trajectories for evaluation and filter out any trajectory that does not reach the goal of each task.

\subsection{Model Details}
\label{sec:app:models}

\textbf{Per-object Encoder CNN.} Conv2D($5{\times}5$ kernel size, 32 kernels, stride 2, padding 1) $\rightarrow$ BatchNorm \citep{ioffe15batch} $\rightarrow$ ReLU $\rightarrow$ Conv2D($5{\times}5$ kernel size, 64 kernels, stride 2, padding 1) $\rightarrow$ BatchNorm $\rightarrow$ ReLU $\rightarrow$ Conv2D($5{\times}5$ kernel size, 64 kernels, stride 2, padding 1).

\textbf{GNN.} Both the node and the edge networks are MLPs with one hidden layer of size 512. Each layer, except the output layer, is followed by LayerNorm \citep{ba16layer} and ReLU activation.

\textbf{Decoder for FWM-AE Baseline.} ConvTranspose2D($5{\times}5$ kernel size, 64 kernels, stride 2) $\rightarrow$ BatchNorm $\rightarrow$ ReLU $\rightarrow$ Conv2D($3{\times}3$ kernel size, 32 kernels, stride 2) $\rightarrow$ BatchNorm $\rightarrow$ ReLU $\rightarrow$ Conv2D($3{\times}3$ kernel size, 14 kernels, stride 1).

\textbf{Training.} We use the Adam \citep{kingma15adam} optimizer with default parameters and a learning rate of $5e-5$. We train for 200 epochs (dataset size is 200k transitions) with a batch size of 256.

\textbf{Planning.} $\delta$, a hyper-parameter of our pick-and-place heuristic, is set to $0.175$ for all tasks. This value was chosen by a grid-search performed on the task of stacking four cubes. All members of an FWM ensemble have identical architectures and training data; they only differ in the inital random seed.

\section{Additional Experiments}
\label{sec:app:additional}

\subsection{C-SWM with Heuristic Action Factorization}

\begin{figure}
    \centering
    \includegraphics[width=1\textwidth]{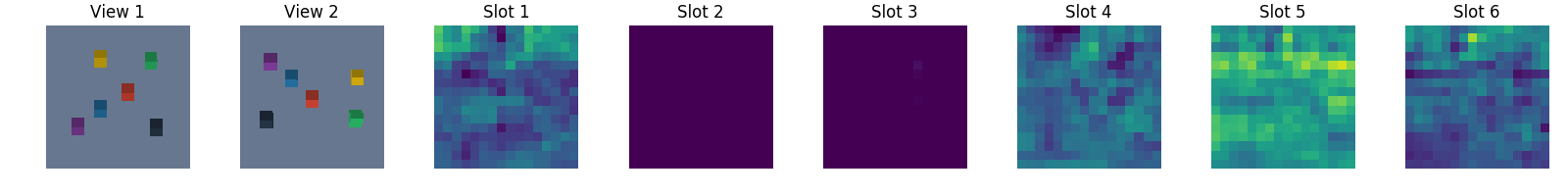} \\
    \includegraphics[width=1\textwidth]{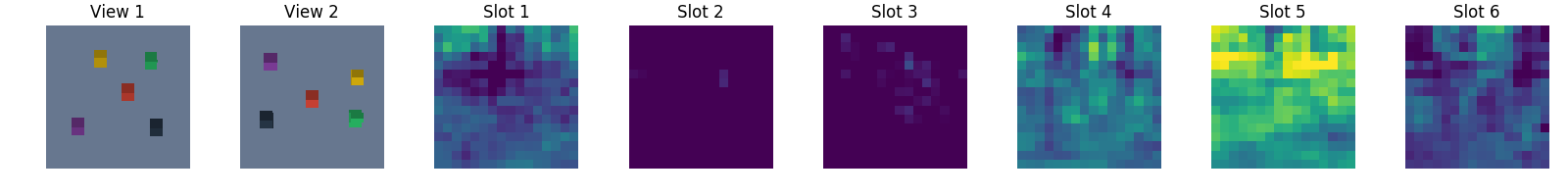} \\
    \includegraphics[width=1\textwidth]{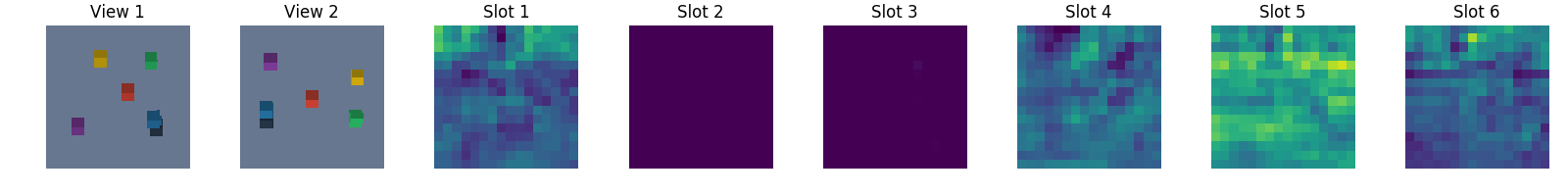} \\
    \includegraphics[width=1\textwidth]{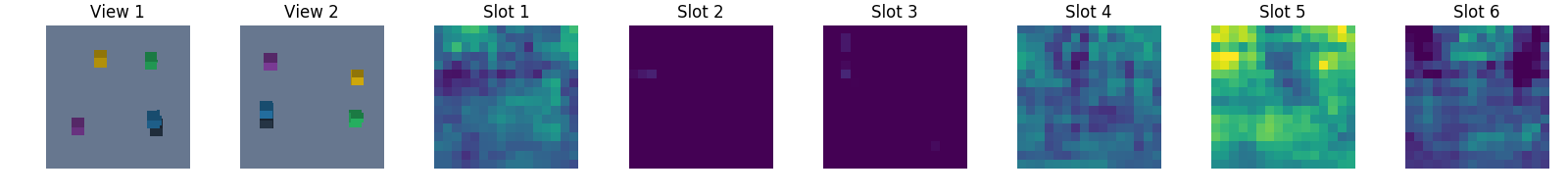} \\
    \includegraphics[width=1\textwidth]{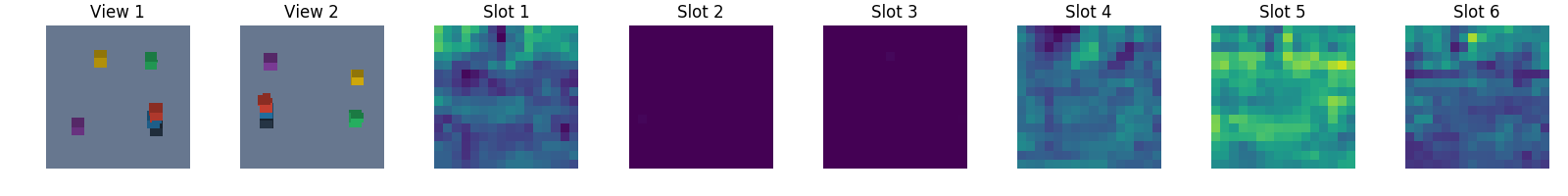} \\
    \includegraphics[width=1\textwidth]{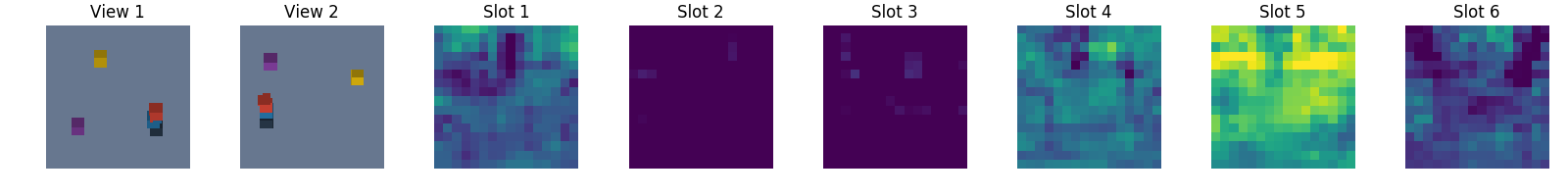} \\
    \includegraphics[width=1\textwidth]{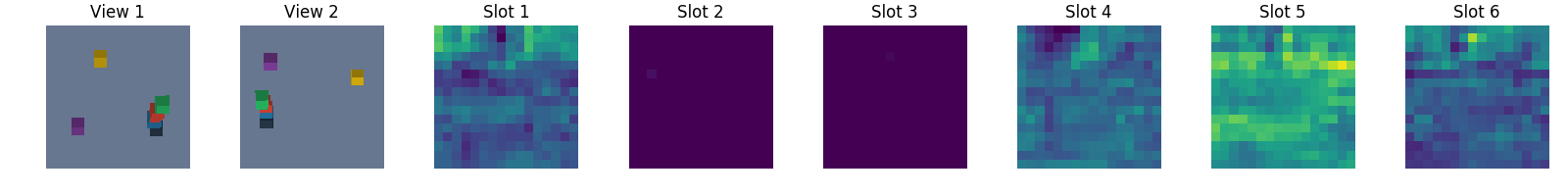}
    \caption{Visualization of feature maps learned by C-SWM with heuristic action factorization. The first two columns show the two views of the environment provided to the model. The next six columns show the $18{\times}18$ feature maps for each object slot given by the C-SWM object extractor. The colormap is scaled between 0 and 0.4.}
    \label{fig:cswm_features}
\end{figure}

We train the C-SWM model \citep{kipf20contrastive} in \textsc{Cubes}. Unlike FWMs, C-SWM does not receive a factored state space; instead, it can choose what information is captured in each object slot. We make two changes to our environment to help C-SWM factor it: (1) we give each object a distinct color (Figure \ref{fig:cswm_features}) so that the model can potentially learn color-specific filters and (2) we create a heuristically factored action space. The factored action space only provides action $a^t_{i}$ to the $i$th node in the C-SWM graph neural network if the $i$th object changed between state $s^t$ and $s^{t+1}$. Otherwise, the $i$th node receives a null action.

The model receives four images concatenated channel-wise: two images of the workspace and two images of the robot hand (which indicate if the robot is holding an object). We use a custom encoder with the following architecture: Conv2D($5{\times}5$ kernel size, 64 kernels, stride 2) $\rightarrow$ BatchNorm $\rightarrow$ LeakyReLU $\rightarrow$ Conv2D($5{\times}5$ kernel size, 64 kernels, stride 2) $\rightarrow$ BatchNorm $\rightarrow$ LeakyReLU $\rightarrow$ Conv2D($5{\times}5$ kernel size, 6 kernels, stride 1) $\rightarrow$ BatchNorm $\rightarrow$ ReLU. The output of the encoder is a $16{\times}16$ feature map for each object.

Across a range of learning rate, C-SWM does not learn to factor the state space of Cube stacking. By our metrics, C-SWM is on-par with an unfactored model (Table \ref{tab:main_comp}). Figure \ref{fig:cswm_features} visualizes the learned feature maps for each object slot: the maps follow an ABAB pattern, where the model appears to only distinguish between the robot holding or not holding an object. This pattern holds across episodes.

\begin{algorithm}
\caption{Pick-and-Place Heuristic for Sequential Tasks}\label{alg:heuristic2} 

\begin{flushleft}
    \hspace*{\algorithmicindent} \textbf{Input:} State embedding $z$, goal embedding $z^g$, action $a$, transition model $g_{\theta}$, in-hand classifier $j_{\psi}$, goal threshold $\delta$. \\
    \hspace*{\algorithmicindent} \textbf{Output:} Heuristic distance. \\
\end{flushleft}

\begin{algorithmic}[1]

    \State Predict next state $\hat{z} = z + g_{\theta}(z, a)$.
    \State Heuristic $h \leftarrow 0$.
    \State Flag $f \leftarrow \text{True}$.
    \For {object $1$ to $K$ in goal sequence}
        \State $d \leftarrow \norm{\hat{z}_k - z^g_k}^2_2$
        \If {$d < \delta$}
            \State $h \leftarrow h + d$. \Comment{Close enough to goal.}
        \ElsIf{$j_{\psi}(\hat{z}_k) = 1$}
            \If {$f$}
                \State $h \leftarrow h + 1$. \Comment{At least one step away from goal.}
            \Else
                \State $h \leftarrow h + 3$. \Comment{Wrong object picked.}
            \EndIf
            \State $f \leftarrow \text{False}$.
        \Else
            \State $h \leftarrow h + 2$. \Comment{At least two steps away from goal.}
            \State $f \leftarrow \text{False}$.
        \EndIf
    \EndFor
    \State \Return $h$.

\end{algorithmic}
\end{algorithm}

\begin{algorithm}
\caption{$L^2$ Heuristic}\label{alg:heuristic_l2} 

\begin{flushleft}
    \hspace*{\algorithmicindent} \textbf{Input:} State embedding $z$, goal embedding $z^g$, action $a$, transition model $g_{\theta}$, in-hand classifier $j_{\psi}$. \\
    \hspace*{\algorithmicindent} \textbf{Output:} Heuristic distance. \\
\end{flushleft}

\begin{algorithmic}[1]

    \State Predict next state $\hat{z} = z + g_{\theta}(z, a)$.
    \State Heuristic $h \leftarrow 0$.
    \State Match objects between $\hat{z}$ and $z^g$ (Section \ref{sec:methods:search}).
    \For {$k$th object}
        \State $d \leftarrow \norm{\hat{z}_k - z^g_k}^2_2$
        \State $h \leftarrow h + d$.
    \EndFor
    \State \Return $h$.

\end{algorithmic}
\end{algorithm}

\end{document}